\documentclass[lettersize,journal]{IEEEtran}
\usepackage{amsmath,amsfonts}
\usepackage{array}
\usepackage{textcomp}
\usepackage{stfloats}
\usepackage{url}
\usepackage{verbatim}
\usepackage{graphicx}

\usepackage{subfigure}
\usepackage{bm}
\usepackage{color}

\usepackage{multirow}
\usepackage[table,xcdraw]{xcolor}
\makeatletter  
\newif\if@restonecol  
\makeatother

\usepackage[linesnumbered,ruled,vlined]{algorithm2e}
\usepackage{algpseudocode}  
\usepackage{amsmath}  

\usepackage{graphicx}
\usepackage{subfigure}
\usepackage{amssymb}

\usepackage{graphicx}                                                           
\usepackage{float} 

\usepackage{titlesec}
\usepackage{afterpage}

\titleformat{\subsubsection}[runin]
{\normalfont\normalsize\bfseries}
{\thesubsubsection}
{1em}
{}
\titleformat{\subsubsection}[runin]
{\normalfont\normalsize\bfseries}
{\thesubsubsection}
{1em}
{}[\newline\hspace*{1em}]
\def\BibTeX{{\rm B\kern-.05em{\sc i\kern-.025em b}\kern-.08em
    T\kern-.1667em\lower.7ex\hbox{E}\kern-.125emX}}
\usepackage{balance}
\begin{document}
\title{Multi-Risk-RRT: An Efficient Motion Planning Algorithm for Robotic Autonomous Luggage Trolley Collection at Airports}

\author{Zhirui Sun$^{\dagger}$, Boshu Lei$^{\dagger}$, Peijia Xie, Fugang Liu, Junjie Gao,\\ Ying Zhang, \emph{Member, IEEE} and Jiankun Wang, \emph{Senior Member, IEEE} 
\thanks{$^{\dagger}$Equal contribution.}%
\thanks{This work is supported by National Natural Science Foundation of China grant \#62103181, \emph{(Corresponding author: Jiankun Wang).}}
\thanks{Zhirui Sun, Peijia Xie and Jiankun Wang are with Shenzhen Key Laboratory of Robotics Perception and Intelligence, Department of Electronic and Electrical Engineering, Southern University of Science and Technology, Shenzhen, China(e-mail: sunzr2023@mail.sustech.edu.cn; xiepj2022@mail.sustech.edu.cn; wangjk@sustech.edu.cn).}%
\thanks{Zhirui Sun, Junjie Gao and Jiankun Wang are with Jiaxing Research Institute, Southern University of Science and Technology, Jiaxing, China  (e-mail: ; junjie.gao1999@gmail.com).}%
\thanks{Boshu Lei is with School of Engineering and Applied Science, University of Pennsylvania, Philadelphia, PA, U.S.A. (e-mail: leiboshu@seas.upenn.edu).}%
\thanks{Fugang Liu is with School of Biomedical Engineering, Shanghai Jiao Tong University, Shanghai, China (e-mail: liufugang@sjtu.edu.cn).}%
\thanks{Ying Zhang is with the School of Electrical Engineering and the Key Laboratory of Intelligent Rehabilitation and Neuromodulation of Hebei Province, Yanshan University, Qinhuangdao, China (e-mail: yzhang@ysu.edu.cn).}%
}


\maketitle

\begin{abstract}
Robots have become increasingly prevalent in dynamic and crowded environments such as airports and shopping malls. In these scenarios, the critical challenges for robot navigation are reliability and timely arrival at predetermined destinations. While existing risk-based motion planning algorithms effectively reduce collision risks with static and dynamic obstacles, there is still a need for significant performance improvements. Specifically, the dynamic environments demand more rapid responses and robust planning. To address this gap, we introduce a novel risk-based multi-directional sampling algorithm, Multi-directional Risk-based Rapidly-exploring Random Tree (Multi-Risk-RRT). Unlike traditional algorithms that solely rely on a rooted tree or double trees for state space exploration, our approach incorporates multiple sub-trees. Each sub-tree independently explores its surrounding environment. At the same time, the primary rooted tree collects the heuristic information from these sub-trees, facilitating rapid progress toward the goal state.
Our evaluations, including simulation and real-world environmental studies, demonstrate that Multi-Risk-RRT outperforms existing unidirectional and bi-directional risk-based algorithms in planning efficiency and robustness.
\end{abstract}


\section{Introduction}
\IEEEPARstart{I}{n} the crowded environment of modern airports, even small inefficiencies such as manual luggage trolley collection can lead to significant delays and operational challenges. As the demand for air travel increases, the potential for robotics to automate this task becomes increasingly appealing \cite{real-time}\cite{xiao-icra}. An efficient and robust motion planning algorithm is essential when introducing robots into these dynamic, human-robot coexistence environments \cite{hr}. Over the past few decades, numerous well-known motion planning methods have been proposed, falling into three categories.
Khatib introduced the artificial potential field method \cite{apf}, which models the robot's movement in the environment as navigation through a virtual artificial force field. However, this method often gets trapped in local minima.
\begin{figure}[t]
\centering
    \includegraphics[width=1\columnwidth]{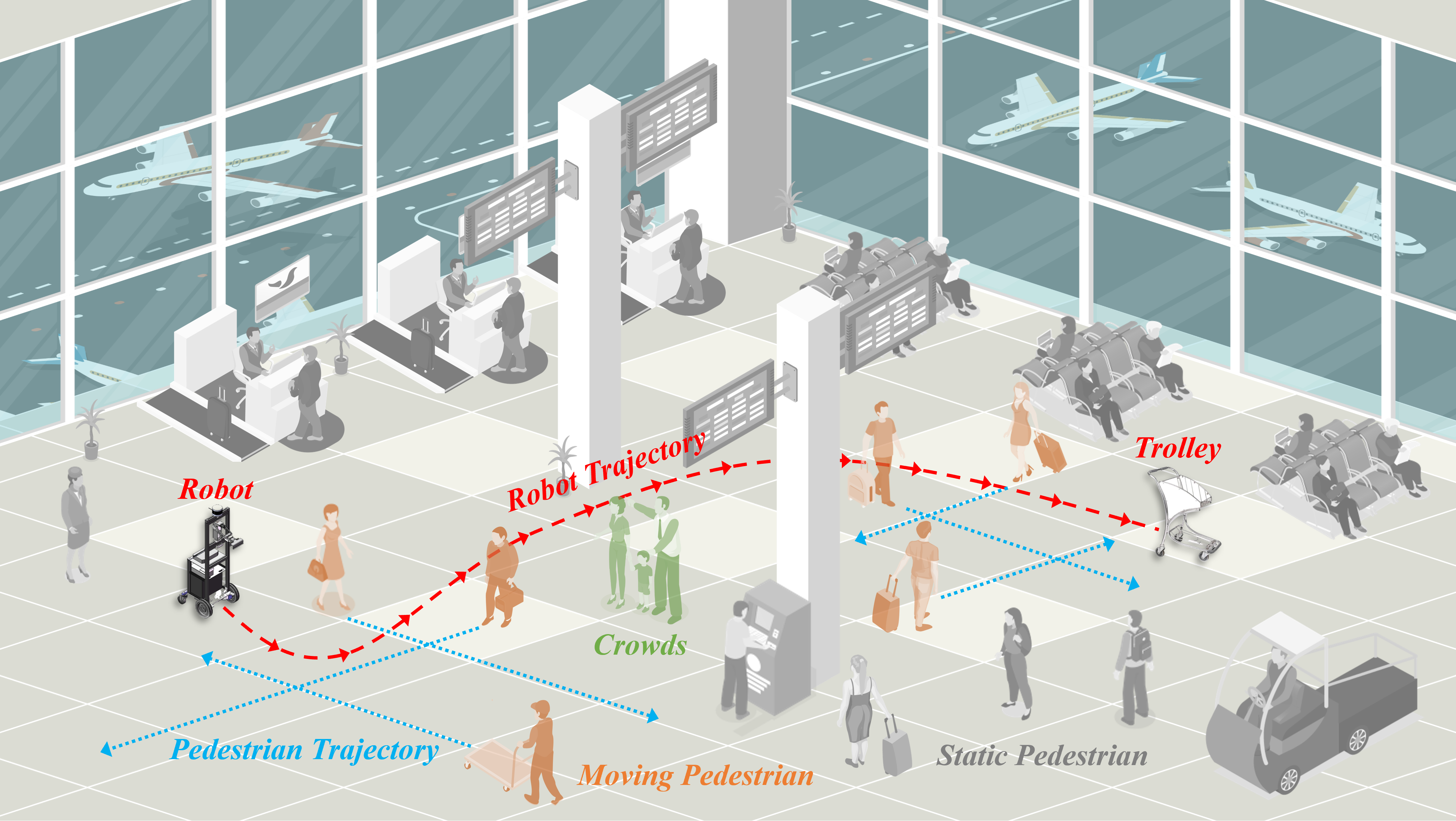}
\caption{Schematic diagram of Multi-Risk-RRT for robotic autonomous luggage trolley collection at airports.}
\label{airport}
\end{figure}
Graph-based planning algorithms, such as A* \cite{A*} and Dijkstra's algorithm \cite{Dijkstra}, utilize graph structures to find optimal trajectories. However, these methods rely on explicit obstacle representations, which can result in significant computational burdens when dealing with high-dimensional environments \cite{computation}.
Sampling-based planning techniques, such as the Rapidly-exploring Random Tree (RRT) algorithm \cite{RRT}, the Probabilistic Roadmap Method (PRM) algorithm \cite{PRM} and the Informed RRT algorithm \cite{Informed_rrt} employ random sampling to expand the search space until a trajectory connecting the start and goal points is found. Unlike graph-based planning algorithms, these methods offer the advantage of quickly producing feasible trajectories without explicit environment representation.

Various approaches have been proposed to improve the performance of sampling-based methods, including bi-directional \cite{rrt-connect} and multi-directional \cite{rrdt} search sampling strategies. However, connecting trees while ensuring adherence to the differential constraints of robot dynamics poses challenges in practical applications. The Two-Point Boundary Value Problem (TBVP) \cite{tbvp} needs to be solved, which is a non-trivial task. Particularly for robots with nonholonomic constraints \cite{constrain}, obtaining reasonable solutions can be challenging and time-consuming in terms of numerical computations. Besides, airports are dynamic, complex spaces with moving passengers and staffs. Traditional motion planning algorithms, often designed for static settings, struggle in dynamic environments. While the Risk-RRT algorithm proposed by Fulgenzi \emph{et al}. \cite{risk-rrt} in 2010 has advanced the field by adapting to dynamic situations, it needs to be further modified and improved to adapt to crowded areas like airports.

This article presents a novel algorithm named Multi-Risk-RRT to overcome the challenges outlined above. A practical demonstration of this algorithm within an airport setting is illustrated in Fig. \ref{airport}. Within the crowded waiting area, the robot employs Multi-Risk-RRT to navigate through the complex moving crowds and static obstacles, ultimately reaching the luggage trolley that needs to be collected. Multi-Risk-RRT aims to enhance planning performance in dynamic and complex environments without relying on TBVP solvers. The contributions of this article are as follows:
\begin{itemize}
    \item This article introduces the Multi-directional Searching and Heuristic Sampling (MSHS) strategy, designed to gather heuristic information from sub-trees, guiding the growth of the rooted tree while bypassing TBVP limitations.
    \item This article presents Multi-Risk-RRT, a novel algorithm that combines the MSHS strategy with a risk-based sampling algorithm.
    \item Multi-Risk-RRT utilizes Heuristic Information Updating, adapted in real-time to dynamic environments, and this feature is validated through comparative experiments.
\end{itemize}

\section{Related Work}
In recent years, several variants of the RRT algorithm have been developed to enhance its search efficiency. Kuffner and LaValle \cite{rrt-connect} introduce the RRT-Connect algorithm, which utilizes two trees originating from the start and goal points to expand bi-directionally. The Learning-Based Multi-RRTs (LM-RRT) approach \cite{lm-rrt} is introduced to solve the challenges of robot motion planning in constrained spaces. Lai \emph{et al}. \cite{rrdt} propose the RRdT* algorithm, which employs multiple local trees to balance exploration and exploitation. Multi-directional search has shown great potential for improving search efficiency. However, these algorithms do not consider the robot's differential and geometry constraints. Moreover, solving the TBVP for tree connection becomes challenging when considering the robot’s kinematic constraints.

Exploring heuristic search methods to avoid the TBVP problem is a promising research direction. Wang \emph{et al}. \cite{b2u} introduce the B2U-RRT algorithm, maintaining two trees growing from the start and goal states, respectively. When these two trees are close, the tree from the goal state serves as heuristic information, guiding the tree's growth from the start state. Sun \emph{et al}. \cite{mt-rrt} propose the MT-RRT algorithm, which utilizes multiple local trees as heuristic information to guide the expansion of the rooted tree. Ichter \emph{et al}. \cite{cave} utilize a Conditional Variational Autoencoder (CVAE) to calculate the sampling distribution. Zhang \emph{et al}. \cite{gan} employ a generative adversarial network (GAN) to generate heuristic regions for achieving non-uniform sampling. 

With the increasing prevalence of human-robot coexistence environments, the scope of robot motion planning has expanded beyond static environments to encompass dynamic and complex settings. Consequently, efficient and robust navigation to reach target locations has become crucial in motion planning. In this context, researchers have developed many algorithms to address these challenges. One such algorithm is RT-RRT*, proposed by Naderi \emph{et al}. \cite{re-rrt}, which introduces an online tree rewiring strategy that allows the tree root to adapt alongside the agent without discarding previously sampled paths. Otte and Frazzoli present RRTX \cite{rrtx}, which maintains the same search graph throughout navigation to handle unpredictable moving obstacles. Zucker \emph{et al}. \cite{mprrt} introduce the Multipartite RRT (MP-RRT), a real-time planning algorithm designed for dynamic environments. It enhances planning efficiency by biasing the sampling distribution and recycling branches from prior planning cycles. These algorithms above rely on frequent replanning or partial modification of their search graphs to respond to environmental shifts. Consequently, these algorithms for replanning must sustain a high update frequency to navigate around moving obstacles effectively. In the case of grid-based methods such as D* \cite{d*} and D*-Lite \cite{d*-lite}, while designed explicitly for dynamic settings, the environment's discretization and complexity highly constrain their effectiveness. 
NR-RRT \cite{nr-rrt} incorporates deep learning techniques using neural network samplers to predict the next promising safe state. However, learning-based approaches may introduce hardware dependency, and achieving real-time performance remains challenging. Chi \emph{et al}. address motion planning in dynamic environments with their Risk-DTRRT algorithm \cite{risk-dtrrt}. Additionally, Ma \emph{et al}. present the Bi-Risk-RRT method \cite{bi-risk-rrt}, where the reverse tree serves as heuristic information instead of explicitly connecting two trees. Although these methods above utilize Risk-RRT for dynamic environments and dual trees for faster growth, they are also limited by these designs. In complex dynamic environments, the efficiency of heuristic information acquisition is reduced, and double trees face challenges with complicated intermediate obstacles, leading to delays in tree connection. To address these issues, we propose to grow multiple RRT trees as sub-trees from multiple directions, making our Multi-Risk-RRT different from previous methods. Utilizing heuristic information acquired from sub-trees, the rooted tree's expansion is enhanced, enabling the robot to navigate through dynamic and complex environments with improved efficiency and robustness.

The subsequent sections of this article are organized as follows. Section III offers the fundamentals of motion planning and associated functions. Section IV provides the details of the Multi-Risk-RRT algorithm. Section V presents and analyzes the simulation and real-world experiment results. Section VI provides a comprehensive discussion of Multi-Risk-RRT's unique features. The final section, Section VII, summarizes key findings and conclusions.

\section{Preliminaries}
This section provides a detailed discussion of three key aspects: robot motion planning, the risk-based sampling algorithm, and the extend function. Each aspect is addressed in separate subsections, namely Sections III-A, III-B, and III-C.

\subsection{Robot Motion Planning Formulation}
In this subsection, we present the formulation of robot motion planning. The state space of robot is denoted as $\mathcal X \in \mathbb{R}^{d}$. At each time step $t$, we define $\mathcal X_{obs}(t) \in \mathcal X$ as the space occupied by obstacles, while $\mathcal X_{free}(t) = \mathcal X \backslash \mathcal X_{obstacle}(t)$ represents the obstacle-free space. The start state and goal state are denoted as $x_{start}\in \mathcal X_{free}(t)$ and $x_{goal} \in \mathcal X_{free}(t)$, respectively. Additionally, the goal region $\mathcal X_{goal}(t)$ is defined as $\mathcal X_{goal}(t) = {x | x\in \mathcal X_{free}(t), ||x-x_{goal}||<r}$, where $r$ is a predefined radius.

In robot motion planning, the control space of robot is denoted as $\mathcal{U} \in \mathbb{R}^{d}$. The objective of motion planning is to find a feasible trajectory $\mathcal{P}$: $[0, T] \rightarrow \mathcal{X}_{free}(t)$, where $\mathcal{P}(0) = x_{start}$ and $\mathcal{P}(T) \in \mathcal{X}_{goal}$. For every $t \in [0, T]$, $\mathcal{P}(t) \in \mathcal{X}_{free}$. The planned trajectory can be realized by determining a sequence of motion controls $u: [0, T] \rightarrow \mathcal{U}$. The state at time $t+1$, denoted as $\mathcal{P}(t+1)$, is determined by the current state $\mathcal{P}(t)$ and the control input $u(t)$. Therefore, the dynamics equation governing the robot motion control \cite{motioncontrol} can be formulated as:
\begin{equation}
\mathcal{P}(t+1) = \mathcal{F}(\mathcal{P}(t), u(t)),
\end{equation}
where $u(t) \in \mathcal{U}$, and $\mathcal{P}(t+1)$ and $\mathcal{P}(t)$ represent two adjacent states.

\subsection{Risk-RRT Algorithm}
The bi-directional and multi-directional approaches primarily differ in their tree growth strategies, but both are based on risk-based sampling algorithms, which are variations of the Risk-RRT algorithm. Therefore, we will first focus on explaining the operational process of the Risk-RRT algorithm, which consists of four main stages:
\begin{enumerate}[]
\item Initialization (lines 1 to 4 of Algorithm 1): All data structures are initialized as empty sets in this phase. The robot has a goal state $x_{goal}$, and the corresponding $Map$ is loaded. The robot's initial position $x_{start}$ is observed and recorded.
\item Trajectory Management (lines 6 to 9 of Algorithm 1): At the beginning of each cycle, the planned trajectory $\mathcal{P}$ is checked and executed. If the trajectory $\mathcal{P}$ is empty, the robot enters a standby state, waiting for a new trajectory $\mathcal{P}$.
\item Node Pruning and Obstacle Prediction (lines 10 to 16 of Algorithm 1): Based on the robot's current state, unreachable nodes are pruned from the tree $\mathcal{T}$. Additionally, the algorithm predicts the positions of the moving obstacles from the current time $t$ to $t+N*\Delta t$.
\item Tree Expansion and Trajectory Selection (lines 17 to 27 of Algorithm 1): The tree $\mathcal{T}$ is expanded in time increments of $\Delta t$, with a depth limit of $N$. In each tree expansion cycle, the node with the highest weight is selected, considering both collision risk and trajectory cost. At the end of each cycle, the robot chooses a new optimal trajectory to follow.
\end{enumerate}

\begin{figure}[t]
\centering
    \includegraphics[width=1\columnwidth]{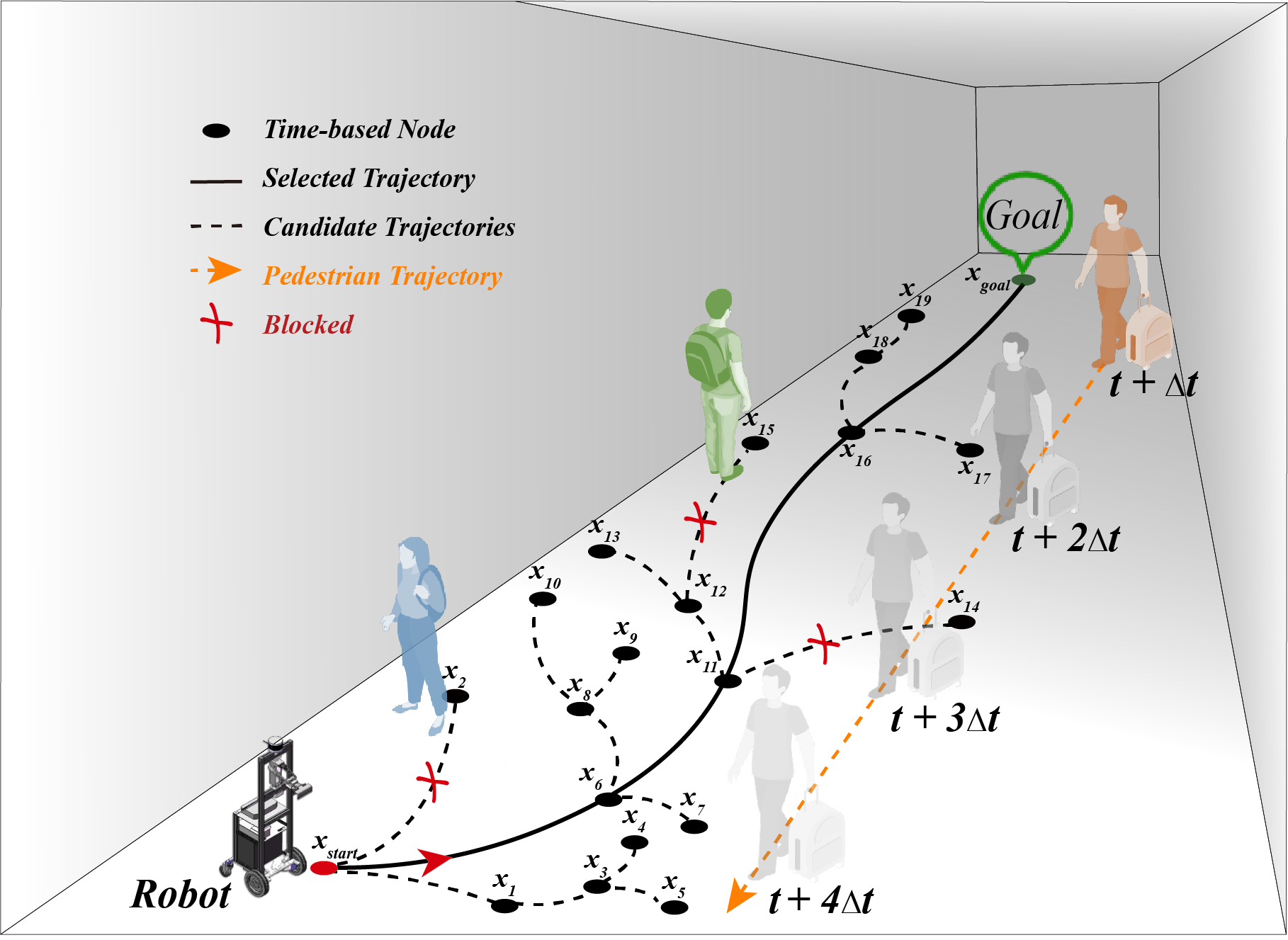}
\caption{Time-based tree in dynamic environment.}
\label{risk}
\end{figure}

These stages collectively form the operational process of the Risk-RRT algorithm, which serves as the foundation for the bi-directional and multi-directional approaches in this study.

In Risk-RRT, the tree is constructed of time-based nodes, thus forming a time-based tree, as illustrated in Fig. \ref{risk}. Specifically, each node, such as $x_1$, $x_2$, etc., originates from an initial state denoted as $x_{start}$. The state of the robot, $x$, includes its position and orientation. The initial and goal states of the robot are represented by $x_{start}$ and $x_{goal}$, respectively. The collision risk probability $P_r(t)$ at a given timestamp $t$, which is influenced by both static and moving obstacles, can be expressed as follows:
\begin{equation} 
P_r(t) = P_{rs} + (1 - P_{rs}) \cdot P_{rd}(t),
\end{equation}
where $P_{rs}$ represents the probability of collision due to static obstacles, and $P_{rd}(t)$ denotes the probability of collision resulting from moving obstacles at timestamp $t$. Furthermore, $P_{rd}(p_i(t))$ represents the probability of collision due to the planned trajectory of a moving obstacle $p_i$ at the same timestamp $t$. The probability of collision from multiple moving obstacles is computed as the complement of the product of their collision probabilities:
\begin{equation}
P_{rd}(t) = 1 - \prod_{i=1}^{n} (1 - P_{rd}(p_i(t))). 
\end{equation}

These equations capture the collision risk probabilities at different timestamps, accounting for both static and moving obstacles and the planned trajectories of the moving obstacles.

\begin{algorithm}  
  \caption{Risk-Based Sampling Algorithm}  
  \KwIn{$x_{start}$, $x_{goal}$, $Map$.}
  \KwOut{$\mathcal P$.}  
  $\mathcal P \gets \emptyset$\;
  $Tree \gets \emptyset$\;
  $x_{state} \gets x_{start}$\;
  $t \gets clock()$\;
  \While {$x_{goal}$ $not$ $reached$}
    {
    \If {$Trajectory$ $is$ $empty$}
        {
            break;
        }
    \Else
        {
            $x_{state} \gets$move along $\mathcal P$ for one step;
        }
    observe($x_{state}$)\;
    delete unreachable trajectories($\mathcal T$, $x_{state}$)\;
    observe($Map$, $movingCrowds$)\;
    $t \gets clock()$\;
    predict $movingCrowds$ state at time $t,...,t+N*\Delta t$\;
    \If {$environment$ $change$}
    {
       update trajectories($Map$, $\mathcal T$, $x_{state}$, $movingCrowds$);
    }
    \While {$clock()<t+\Delta t$}
    {
    \If{Multi-Risk-RRT}
    {
    $multiGrow(\mathcal T, x_{goal}, N, Map)$\;
    }
    \ElseIf{Bi-Risk-RRT}
    {
    $biGrow(\mathcal T, x_{goal}, N, Map)$\;
    }
    \Else
    {
    $x_{rand} \gets uniformSample(Map)$\;
    $x_{new} \gets Extend(\mathcal T, x_{rand}, N)$\;
    \If {$x_{new} \in Regin(x_{goal})$}
             {
                \Return $Reached$\;
             }
    }
    }
    $\mathcal P$ = Choose best trajectory in $\mathcal T$\;
    $t \gets clock()$\;
} 
\end{algorithm} 

\subsection{Extend Function}
In this article, the cost between two states, $x_1$ and $x_2$, is defined as follows:
\begin{equation}
\operatorname{Cost}=w_1 \frac{|| x_1-x_2 \|}{\left\|x_1-x_{goal}\right\|}+w_2 \arccos \left(\frac{\overrightarrow{v_1} \cdot \overrightarrow{x_1 x_2}}{\left|\overrightarrow{v_1}\right|\left|\overrightarrow{x_1 x_2}\right|}\right),
\end{equation}
where $w_1$ and $w_2$ are pre-established constants used to harmonize the impacts of positional and orientational differences. $\overrightarrow{v_1}$ represents the robot's velocity vector in $x_1$ and $L2$ norm is denoted by $||\cdot||$.

Algorithm 2 presents the Extend function, which is crucial in the risk-based sampling algorithm. Given a state $x_{rand}$, we can calculate the cost value $\mathcal C_{k}$ between $x_{rand}$ and $x_k$. This cost value allows us to determine the optimal node $x_{best}$ from the tree $\mathcal{T}$. By examining $x_{best}$, we can extract its current velocity $\upsilon_{current}$ and angular velocity $\omega_{current}$. With the knowledge of acceleration parameters $\alpha_\upsilon$ and $\alpha_\omega$, we can establish the velocity range $[\upsilon_{min}, \upsilon_{max}]$ and angular velocity range $[\omega_{min}, \omega_{max}]$ for a specific time interval referred to as $TimeStep$.

We utilize their respective discrete fractions, $\delta_{n\upsilon}$ and $\delta_{n\omega}$ to discretize the velocity and angular velocity ranges. We can generate a set of expected states $x_{expected}$ by considering different combinations of linear and angular velocities. Next, we evaluate the cost value $\mathcal C_{2}$ between $x_{rand}$ and $x_{expected}$. Using this cost value, we employ the function $chooseBestControl$ to obtain a new node $x_{new}$. If the depth of $x_{new}$ falls within the specified limit of $N$ and the node is unoccupied, we add $x_{new}$ to the tree $\mathcal{T}$. This process enables the expansion and growth of the tree structure.

\begin{algorithm}  
  \caption{Extend Function}  
  \SetKwFunction{Extend}{Extend}
  \SetKwProg{Fn}{Function}{:}{}
    \Fn {$Extend(\mathcal T, x_{rand}, N)$}
    {
    \For{$i=1,...,\mathcal T.size()$}
        {
        $x_k \gets GetNode(\mathcal T)$\;
        $\mathcal C_{k} \gets$ Cost($x_k, x_{rand}$)\;
        $\mathcal C_{node} \gets \frac{1}{\mathcal C_{k} + \beta*x_k.risk}$\;
        $\mathcal C_1 \gets \{\mathcal C_{node}\}$\;
        }
    
    $x_{best} \gets chooseBestNode(\mathcal C_1)$\;

    $\upsilon_{current}, \omega_{current} \gets CurrentState(x_{best})$\;
    $\upsilon_{max}, \upsilon_{min} \gets TimeStep(\upsilon_{current}, \alpha_{\upsilon})$\;
    $\omega_{max}, \omega_{min} \gets TimeStep(\omega_{current}, \alpha_{\omega})$\;
    $\delta_\upsilon \gets \Delta(\upsilon_{max}, \upsilon_{min}, \delta_{n\upsilon})$\;
    $\delta_\omega \gets \Delta(\omega_{max}, \omega_{min}, \delta_{n\omega})$\;
    \For{$i=0$; $i \leq \delta_{n\upsilon}$; $i++$ }
        {
        \For{$j=0$; $j \leq \delta_{n\omega}$; $j++$ }
            {
             $x_{expected} \gets Control(\delta_\upsilon, \delta_\omega, i, j)$\;
             $\mathcal C_{k} \gets$ Cost($x_{expected}, x_{rand}$)\;
             $\mathcal C_2 \gets \{\mathcal C_{k}\}$\;
            }
        }
    $x_{new} \gets chooseBestControl(\mathcal C_2)$\;
    \If{$x_{new}.depth \leq N$ and $x_{new}.isFree$}
    {
        $\mathcal{T} \gets AddNode(x_{new})$\;
        \Return $x_{new}$\;
    }
    }
\end{algorithm} 

\section{Algorithms}
This section will explain the Multi-Risk-RRT algorithm and highlight the differences between the bi-directional and multi-directional approaches. The fundamental idea behind Multi-Risk-RRT is to utilize multiple sub-trees as heuristic information sources, guiding the rooted tree's expansion toward the goal region. The section is organized as follows: Subsection IV-A describes the Bi-Risk-RRT grow function, providing a foundation for comparison. Subsection IV-B shifts the focus to elaborate on the Multi-Risk-RRT grow function, stressing its unique attributes and advantages. Subsection IV-C provides a visual description to illuminate the search process of the Multi-Risk-RRT algorithm.
\subsection{Bi-Risk-RRT Grow Function}
\begin{algorithm}  
    \caption{Bi-Risk-RRT Grow Function}  
  
\SetKwFunction{Extend}{Extend}
\SetKwProg{Fn}{Function}{:}{}
\Fn {$biGrow(\mathcal T, x_{goal}, N, Map)$}
{
$\mathcal V_{goal} \gets \{x_{goal}\}, \mathcal E_{goal} \gets \emptyset,$
$\tau_{goal} \gets (\mathcal V_{goal}, \mathcal E_{goal})$\; 
\If{$meet(\mathcal T, \tau_{goal}) == FALSE$}
    {
        $x_{rand} \gets uniformSample(Map)$\;
        
        $x_{new} \gets Extend(\mathcal T, x_{rand}, N)$\;
        $x_{new} \gets Extend(\tau_{goal}, x_{rand}, N)$\;

    }
    \Else
    {
           
        $x_{rand} \gets heuristicSample(\tau_{goal})$\;
        $x_{new} \gets Extend(\mathcal T, x_{rand}, N)$\;
            \If {$x_{new} \in Regin(x_{goal})$}
             {
                \Return $Reached$\;
             }
    }
    }
\end{algorithm} 

The Bi-Risk-RRT algorithm ( Algorithm 3) follows a similar procedure to Risk-RRT but adds two trees growing simultaneously from the start state $x_{start}$ and the goal state $x_{goal}$. These trees expand toward each other until they intersect. The tree originating from $x_{goal}$ stops growing at the intersection point and serves as a heuristic trajectory from $x_{goal}$ to the intersection point. The nodes along this trajectory form a heuristic sampling distribution, which improves the rooted tree's growth efficiency by guiding it toward potentially feasible solutions.

\subsection{Multi-Risk-RRT Grow Function}
In contrast to unidirectional and bi-directional search methods, the Multi-Risk-RRT algorithm maintains multiple trees exploring the state space: a rooted tree $\mathcal{T}$ originating from the start state $x_{start}$ and sub-trees $\mathcal T\{\tau_{info}\}$, randomly generated via $RandomGenerate$ within the state space $Map$. Each tree investigates local environments independently. When $\mathcal{T}$ is sufficiently close to a $\tau_{info}$, which is a member of $\mathcal T\{\tau_{info}\}$, $\tau_{info}$ functions as a heuristic guide to direct $\mathcal{T}$. Over time, $\mathcal{T}$ accumulates information from $\mathcal T\{\tau_{info}\}$, facilitating rapid growth toward the goal state. During the initial phase, our approach involves uniformly sampling across the map:
\begin{equation}
x_{\text {rand }} \sim \mathbb{U}( { Map }),
\end{equation}
where $\mathbb{U}$ denotes the uniform distribution. In the subsequent phase, we integrate the heuristic distribution with the uniform distribution, ensuring the maintenance of probability completeness throughout the process.
\begin{equation}
x_{\text {rand }} \sim \frac{h_r}{L} \sum_{\kappa=0}^{L-1} \mathbb{N}\left(\mu_\kappa, \sigma_\kappa^2\right)+\left(1-h_r\right) \mathbb{U}( { Map }),
\end{equation}
where $L$ denotes the number of nodes within the heuristic trajectory, while $\mathbb{N}\left(\mu_\kappa, \sigma_\kappa^2\right)$ signifies a Gaussian distribution with a mean of $\mu_\kappa$ and a standard deviation of $\sigma_\kappa$, corresponding to each node $n_i$ within the heuristic trajectory. The value of $\sigma_\kappa$ can be appropriately determined based on the map's dimensions.

We will introduce two foundational aspects of the Multi-Risk-RRT algorithm: the MSHS strategy and Heuristic Information Updating. These two innovative parts significantly optimize the algorithm's efficiency and robustness in high-dynamic and complex settings.

\subsubsection*{MSHS Strategy}
\emph{1) Multi-directional Searching (Line 5-15, Algorithm 4):} In this phase, the state $x_{rand}$ obtained from $uniformSample$ is added to $\mathcal{T}$ or any sub-tree within $\mathcal T\{\tau_{info}\}$. If the distance $Dist(x_{rand},\mathcal{T})$ is less than $\lambda$, $x_{rand}$ is incorporated into $\mathcal{T}$ via the $Extend$ function. Similarly, if $Dist(x_{rand}, \mathcal T\{\mathcal \tau_{info}\}) < \lambda$, $x_{rand}$ is added to the corresponding sub-tree through $AddNode$. Otherwise, a new sub-tree $\tau_{info}$ is generated at $x_{rand}$ via $RandomGenerate$ and subsequently added to $\mathcal T\{\tau_{info}\}$ using $AddTree$.

\emph{2) Heuristic Sampling (Line 16-27, Algorithm 4):} During this phase, if two trees are in close proximity, they are merged into a single entity. Two trees ($\tau_{tree1},\tau_{tree2}$) can be obtained through the tree connection function ($meet$). If one of these trees ($\tau_{tree1}$) is $\mathcal T$, the other tree ($\tau_{tree}$) serves as heuristic information to guide $\mathcal T$ in obtaining the biased sampling state $x_{rand}$ via $heuristicSample$. Subsequently, a new state $x_{new}$ can be extended to $\mathcal T$ via the $Extend$ function. If not, the two neighboring nodes are linked, and all nodes from one sub-tree ($\tau_{tree2}$) are integrated into the other sub-tree ($\tau_{tree1}$).

\subsubsection*{Heuristic Information Updating}
Updating heuristic information manifests in two principal parts (Line 22 and 27, Algorithm 4). Firstly, upon utilization of a sub-tree by $\mathcal T$, the corresponding sub-tree $\tau_{tree}$ is eliminated from the search space via $Delete$ function. This step clears consumed heuristic information, making room for new sub-trees. Secondly, when two sub-trees are close, they are merged into a single sub-tree $\tau_{tree1}$, and the other sub-tree $\tau_{tree2}$ will be removed from the state space using $Delete$ function. This consolidation effectively reduces the computational burden of managing multiple sub-trees, improving the algorithm's operational efficiency.

\begin{algorithm}  
  \caption{Multi-Risk-RRT Grow Function}  
\SetKwFunction{Extend}{Extend}
\SetKwProg{Fn}{Function}{:}{}
\Fn {$multiGrow(\mathcal T, x_{goal}, N, Map)$}
{

$\mathcal V_{goal} \gets \{x_{goal}\}, \mathcal E_{goal} \gets \emptyset,$
$\tau_{goal} \gets (\mathcal V_{goal}, \mathcal E_{goal})$\; 

$\mathcal T\{\tau_{info}\} \gets AddTree(\tau_{goal})$\;


\While {$n \leq N$}
{
    \If{$meet(\mathcal T, \mathcal T \{\tau_{info}\}) == FALSE$}
    {
        $x_{rand} \gets uniformSample(Map)$\;
        \If {$Dist(x_{rand}, \mathcal T) < \lambda$}
        {
            $x_{new} \gets Extend(\mathcal T, x_{rand}, N)$\;
            \If {$x_{new} \in Regin(x_{goal})$}
                     {
                        \Return $Reached$\;
                     }
        }
        \ElseIf {$Dist(x_{rand}, \mathcal T\{\mathcal \tau_{info}\}) < \lambda$}
        {
            
            $\tau_{info} \gets AddNode(x_{rand})$\;
        
        }
        \Else
        {
            $\tau_{info} \gets RandomGenerate(Map)$\;
            $\mathcal T\{\tau_{info}\} \gets AddTree(\tau_{info})$\;
        }
    }
    \If{$meet(\mathcal T, \mathcal T\{\tau_{info}\}) == TRUE$}
    {
        $\tau_{tree1},\tau_{tree2} \gets meet(\mathcal T, \mathcal T\{\tau_{info}\})$\;
            \If{$\{\tau_{tree1}, \tau_{tree2}\} \cap \mathcal T  ==\mathcal T$}
            {
                $\tau_{tree} \gets \{\tau_{tree1}, \tau_{tree2}\}\setminus \mathcal T$\;
                $x_{rand} \gets heuristicSample(\tau_{tree})$\;
                $x_{new} \gets Extend(\mathcal T, x_{rand}, N)$\;
                $Delete(\tau_{tree})$\; 
                    \If {$x_{new} \in Regin(x_{goal})$}
                     {
                        \Return $Reached$\;
                     }
            }
            \Else
            {
                $\tau_{tree1} \gets ExtendTree(\tau_{tree1},\tau_{tree2})$\;
                $Delete(\tau_{tree2})$\;
            }
    }
}
}
\end{algorithm}

\subsection{Multi-Risk-RRT Search Process}
\begin{figure}[t]
\centering
    \subfigure[t = 3.4 s.]{\includegraphics[width=0.49\columnwidth]{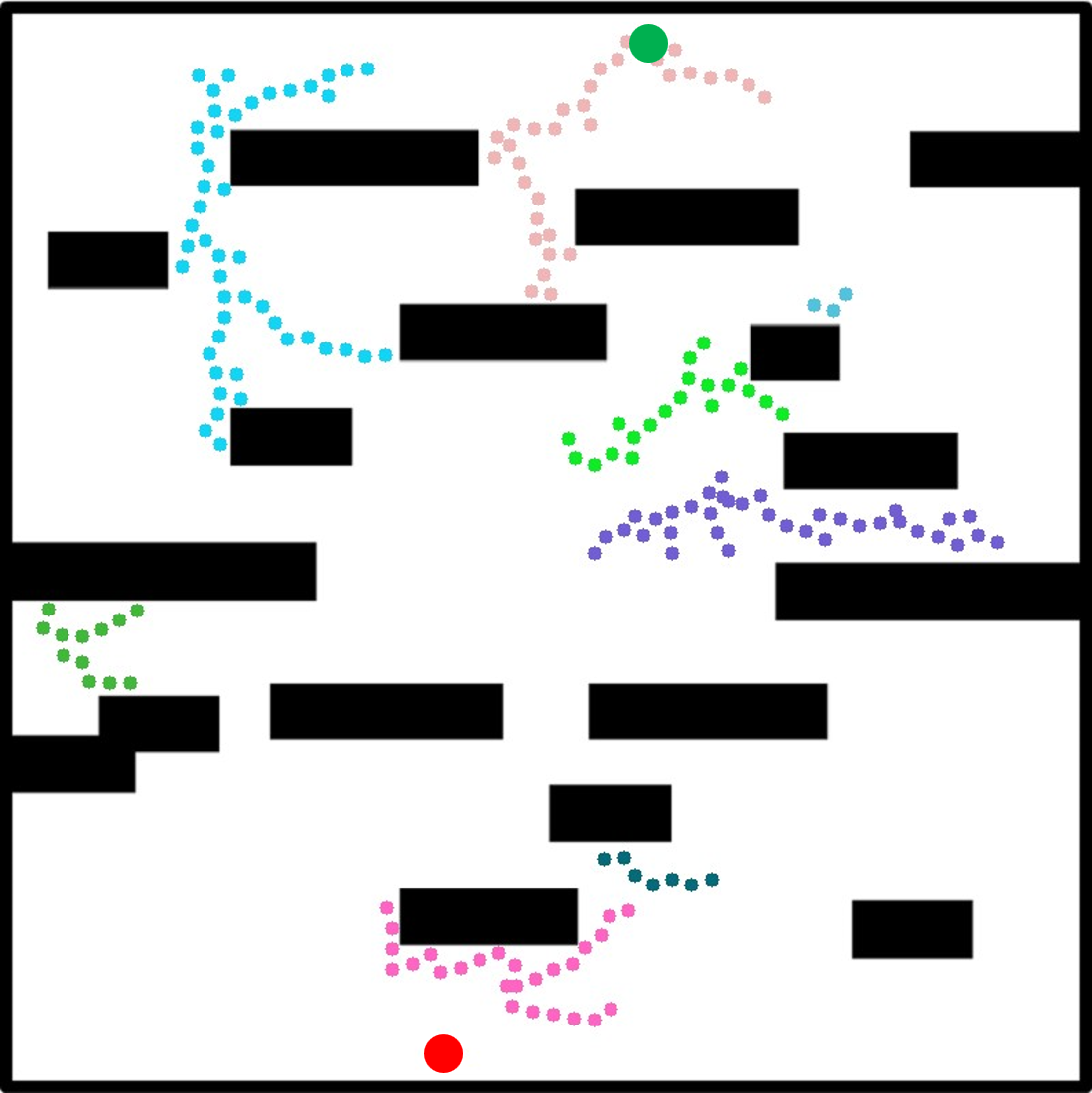}}
    \subfigure[t = 9.5 s.]{\includegraphics[width=0.49\columnwidth]{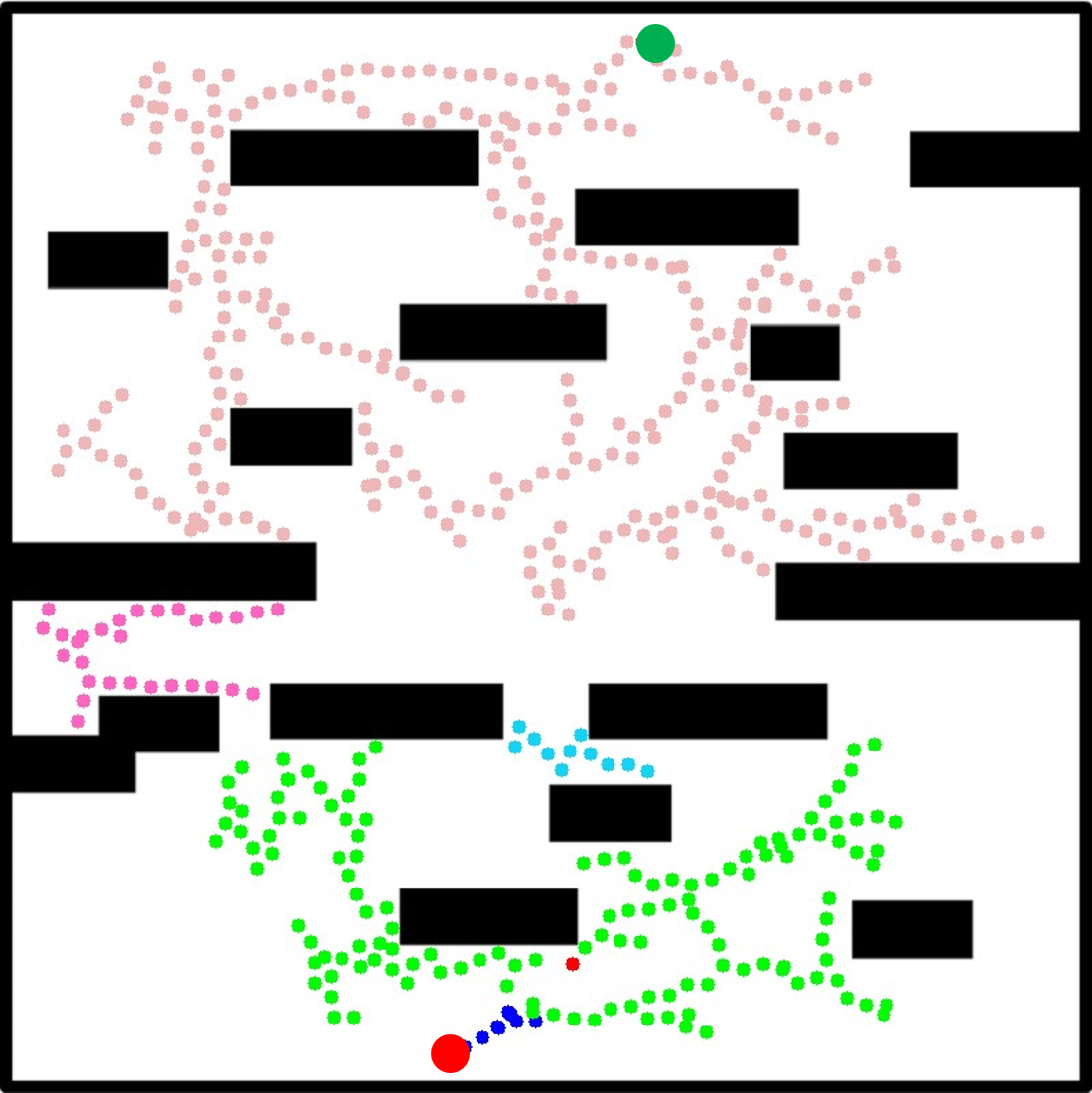}}
    \subfigure[t = 15.6 s.]{\includegraphics[width=0.49\columnwidth]{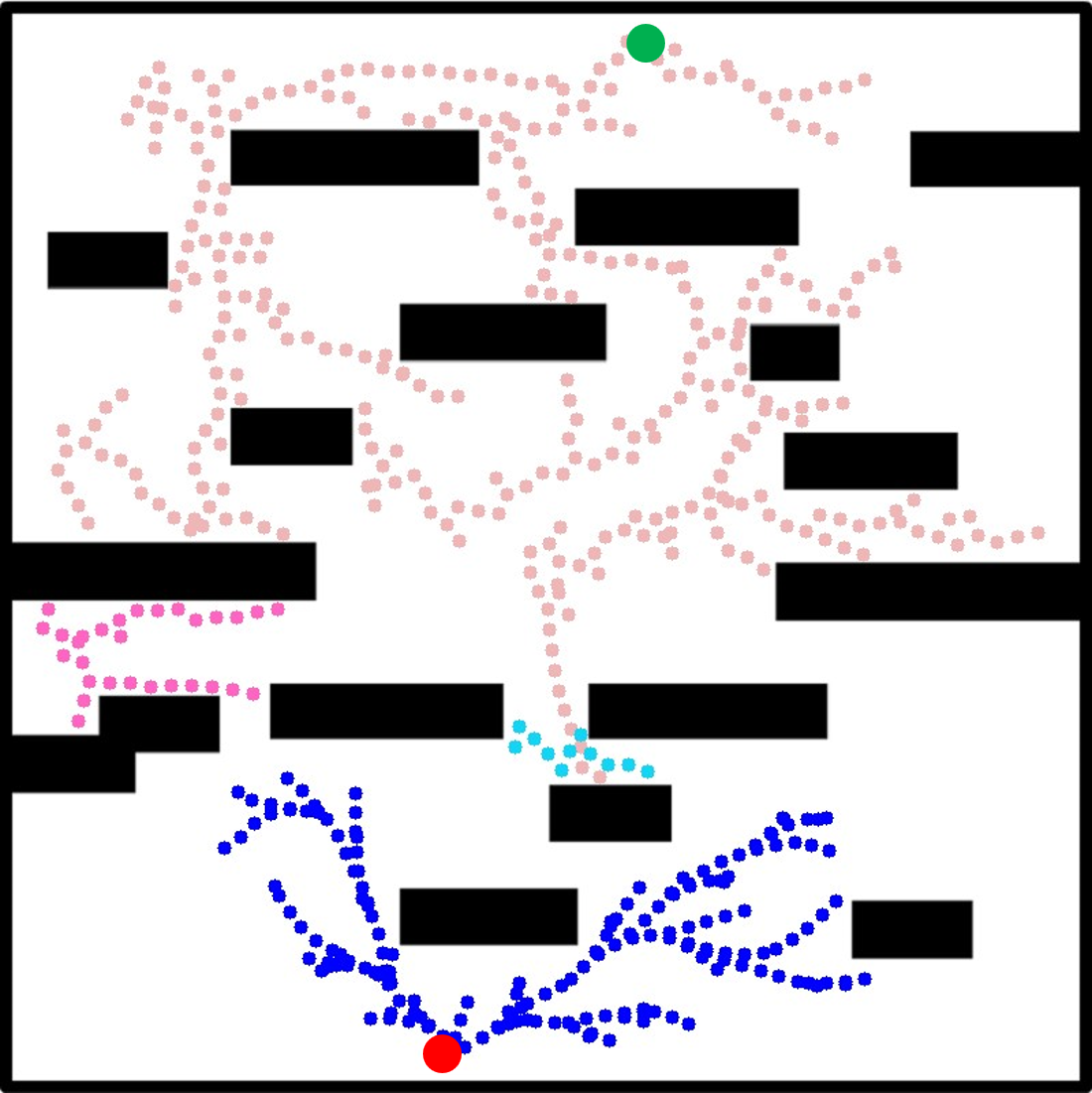}}
    \subfigure[t = 19.5 s.]{\includegraphics[width=0.49\columnwidth]{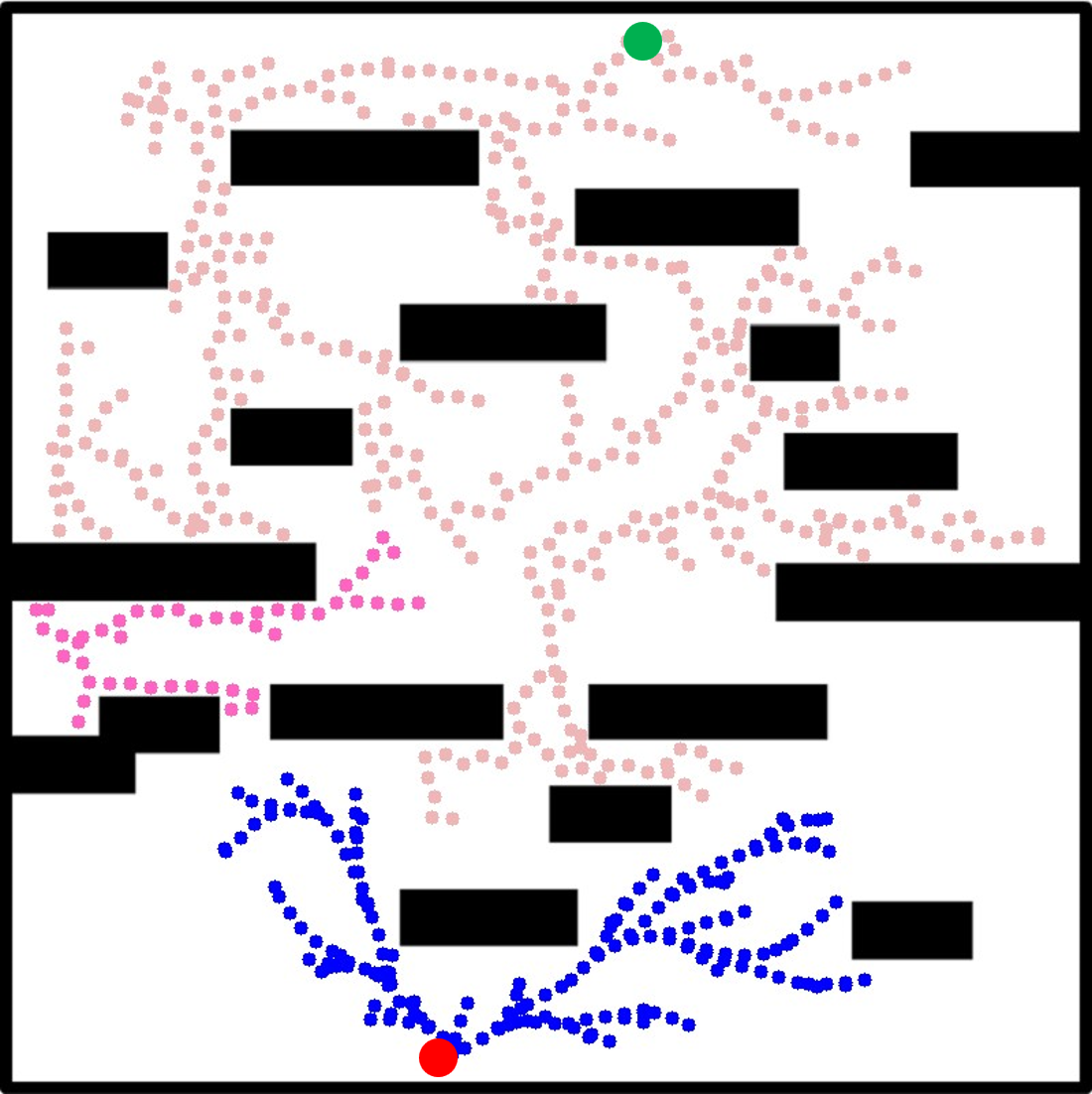}}
    \subfigure[t = 21.1 s.]{\includegraphics[width=0.49\columnwidth]{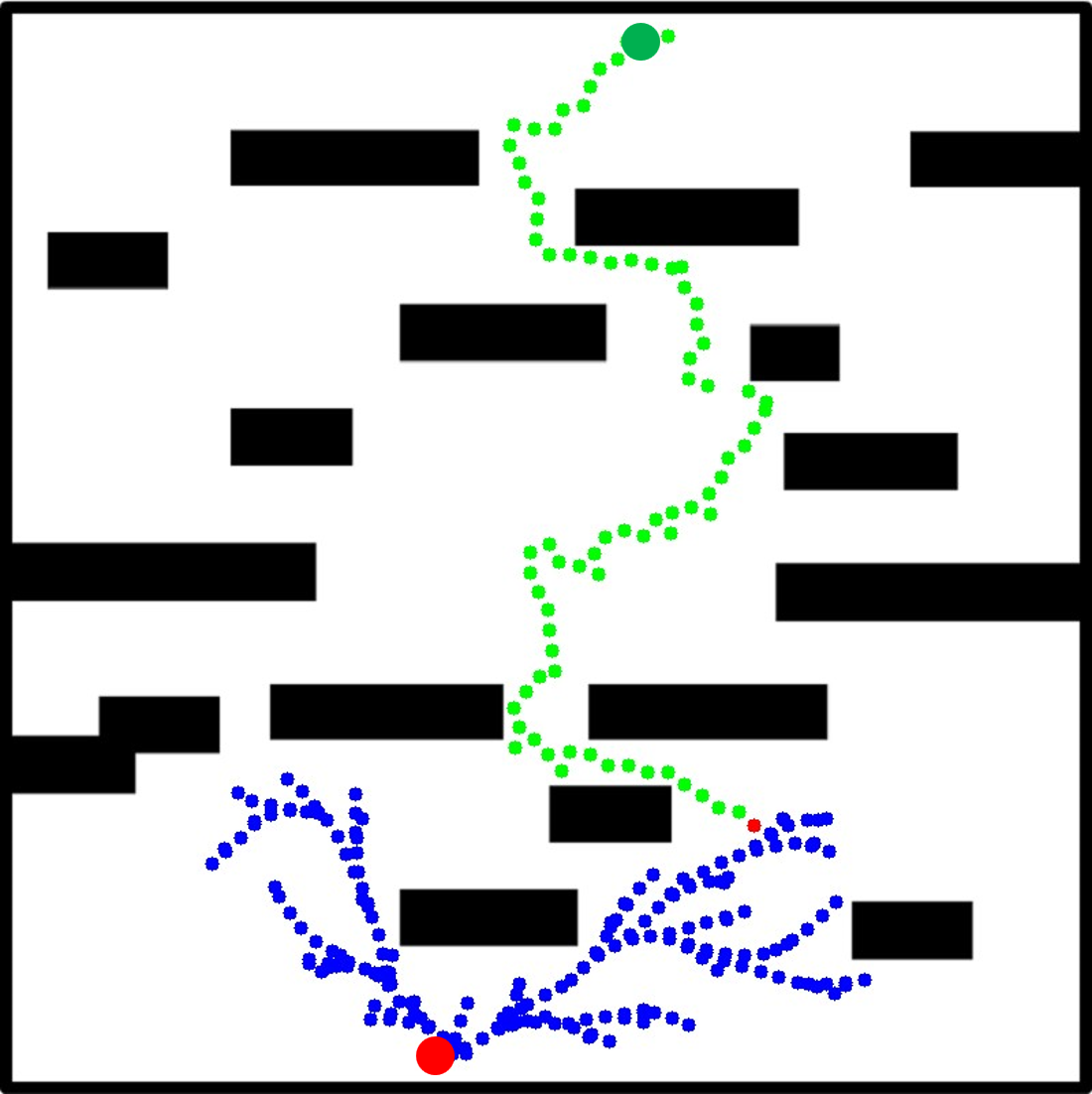}}
    \subfigure[t = 43.7 s.]{\includegraphics[width=0.49\columnwidth]{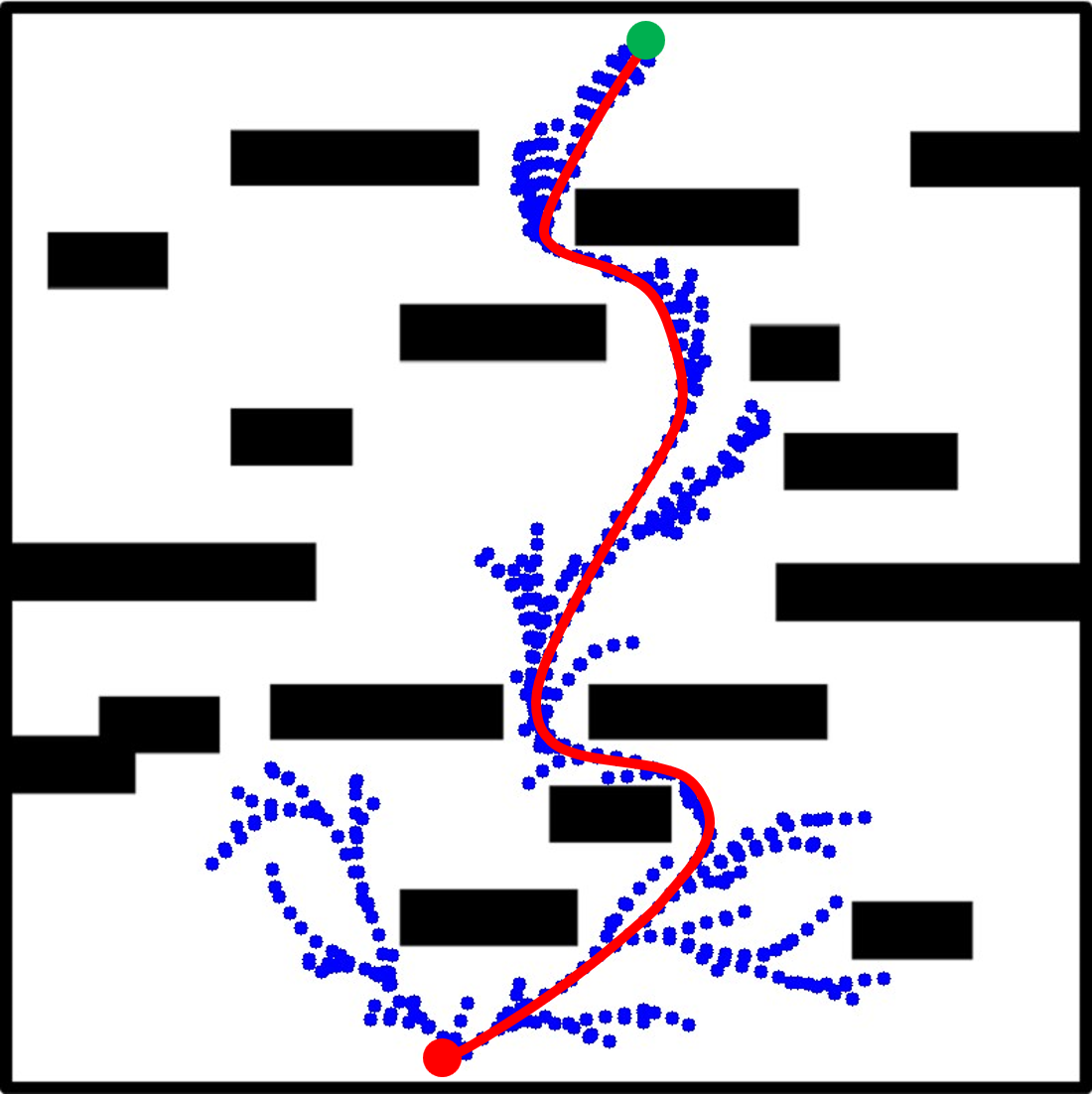}}
\caption{The search process of Multi-Risk-RRT at different crucial stages. (a) t = 3.4 s, (b) t = 9.5 s, (c) t = 15.6 s, (d) t = 19.5 s, (e) t = 21.1 s, and (f) t = 43.7 s.}
\label{Multi-Risk-RRT}
\end{figure}

To provide a more comprehensive illustration of the search process of Multi-Risk-RRT, we highlight six key stages of its operation on Map \uppercase\expandafter{\romannumeral1}, as shown in Fig. \ref{Multi-Risk-RRT}. At t = 3.4 s (see Fig. \ref{Multi-Risk-RRT}(a)), the tree expands incrementally and randomly within the space. When nodes from separate sub-trees find themselves in close spatial proximity, these sub-trees automatically merge. Furthermore, if nodes from any sub-tree are in the surroundings of the rooted tree, they are computationally modeled using a Gaussian Mixture Model (GMM) \cite{gmm} to act as heuristic information to guide the growth of the rooted tree (see Fig. \ref{Multi-Risk-RRT}(b) at t = 9.5 s and Fig. \ref{Multi-Risk-RRT}(c) at t = 15.6 s).
As the algorithm proceeds, these sub-trees merge and expand, eventually forming one or several significant sub-trees, as evident in Fig. \ref{Multi-Risk-RRT}(d) at t = 19.5 s. Importantly, if a node from a sub-tree containing the goal point comes into proximity with a node from the rooted tree, only the heuristic nodes leading from the goal point to the rooted tree node is retained, while other nodes are pruned (see Fig. \ref{Multi-Risk-RRT}(e) at t = 21.1 s). 
Ultimately, as shown in Fig. \ref{Multi-Risk-RRT}(f) at t = 43.7 s, the rooted tree effectively utilizes this heuristic guidance to aggressively grow in the most promising direction, successfully finding a feasible trajectory from the start point to the goal point.

\section{Experiments}
\begin{figure}[t]
\centering
    \subfigure[Map \uppercase\expandafter{\romannumeral1}.]{\includegraphics[width=0.49\columnwidth]{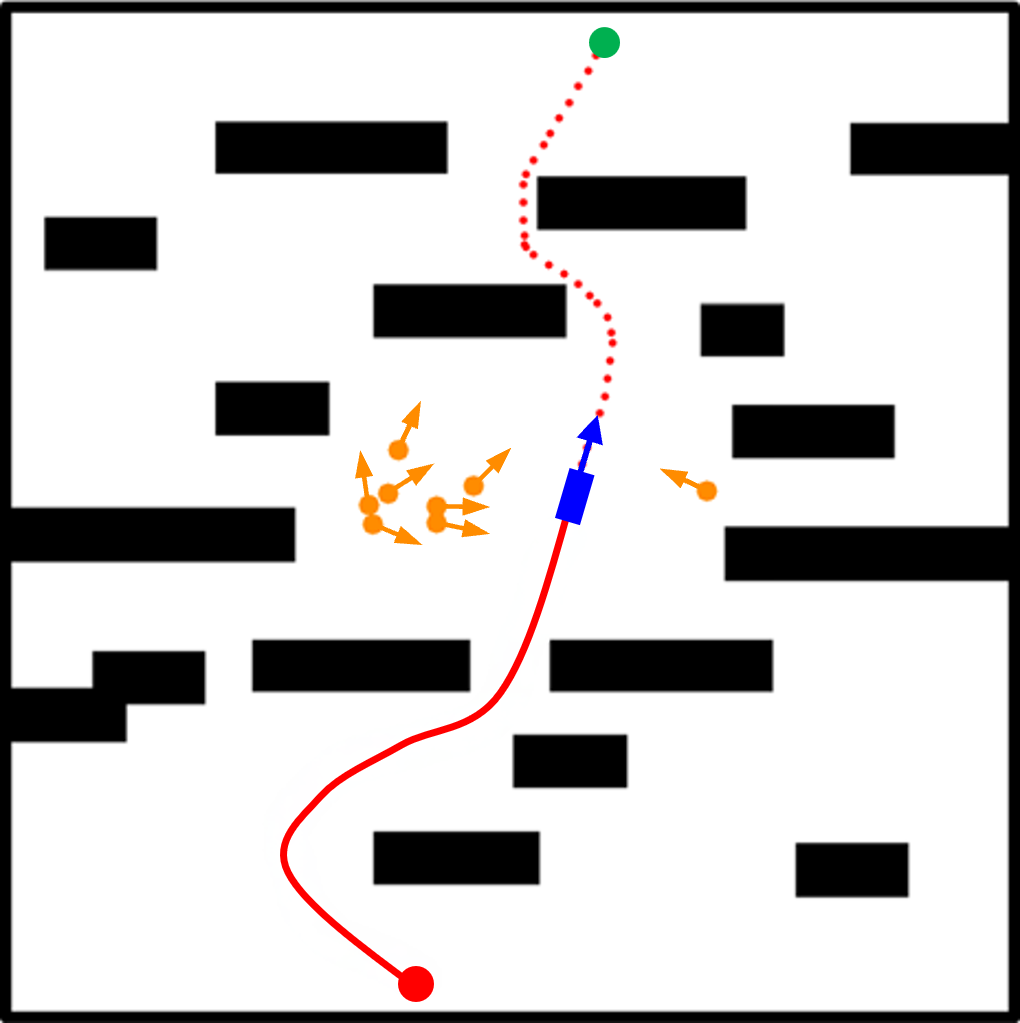}}
    \subfigure[Map \uppercase\expandafter{\romannumeral2}.]{\includegraphics[width=0.49\columnwidth]{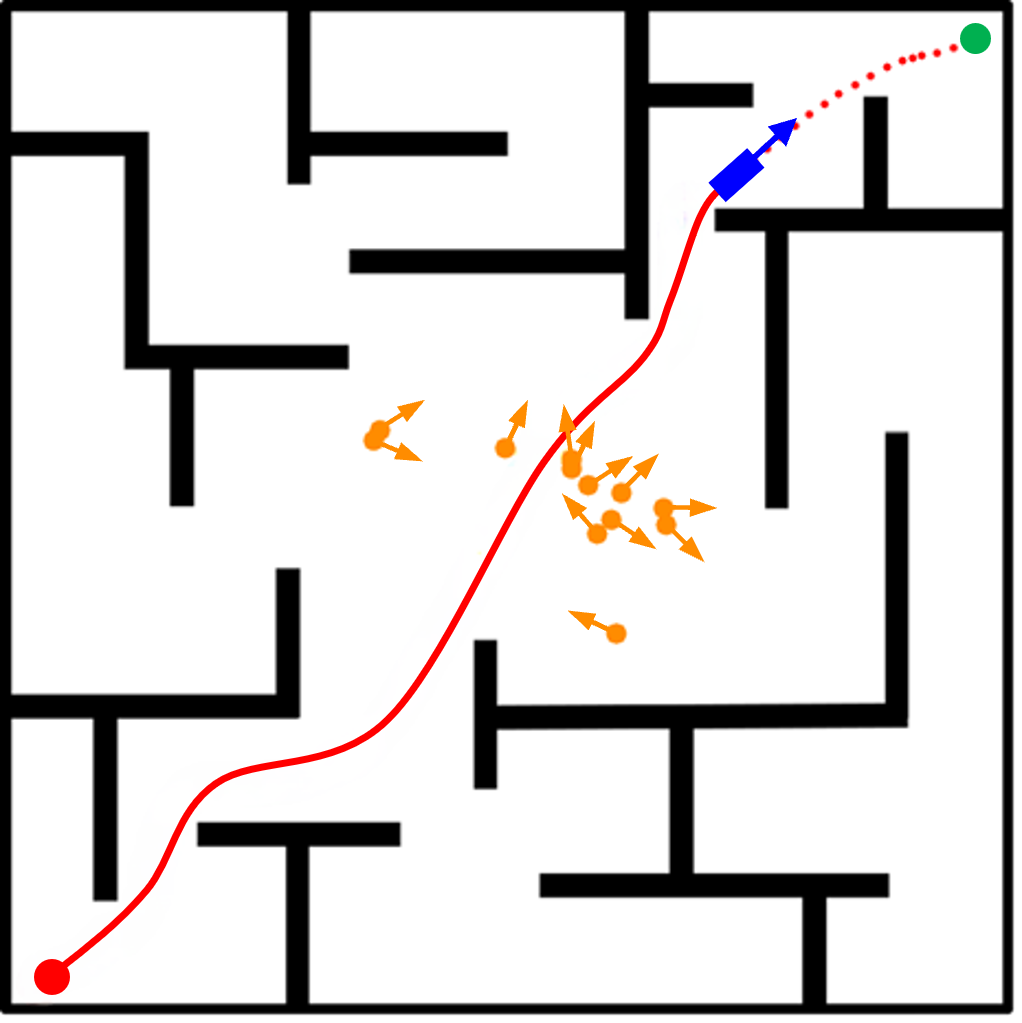}}
    \subfigure[Map \uppercase\expandafter{\romannumeral3}.]{\includegraphics[width=0.49\columnwidth]{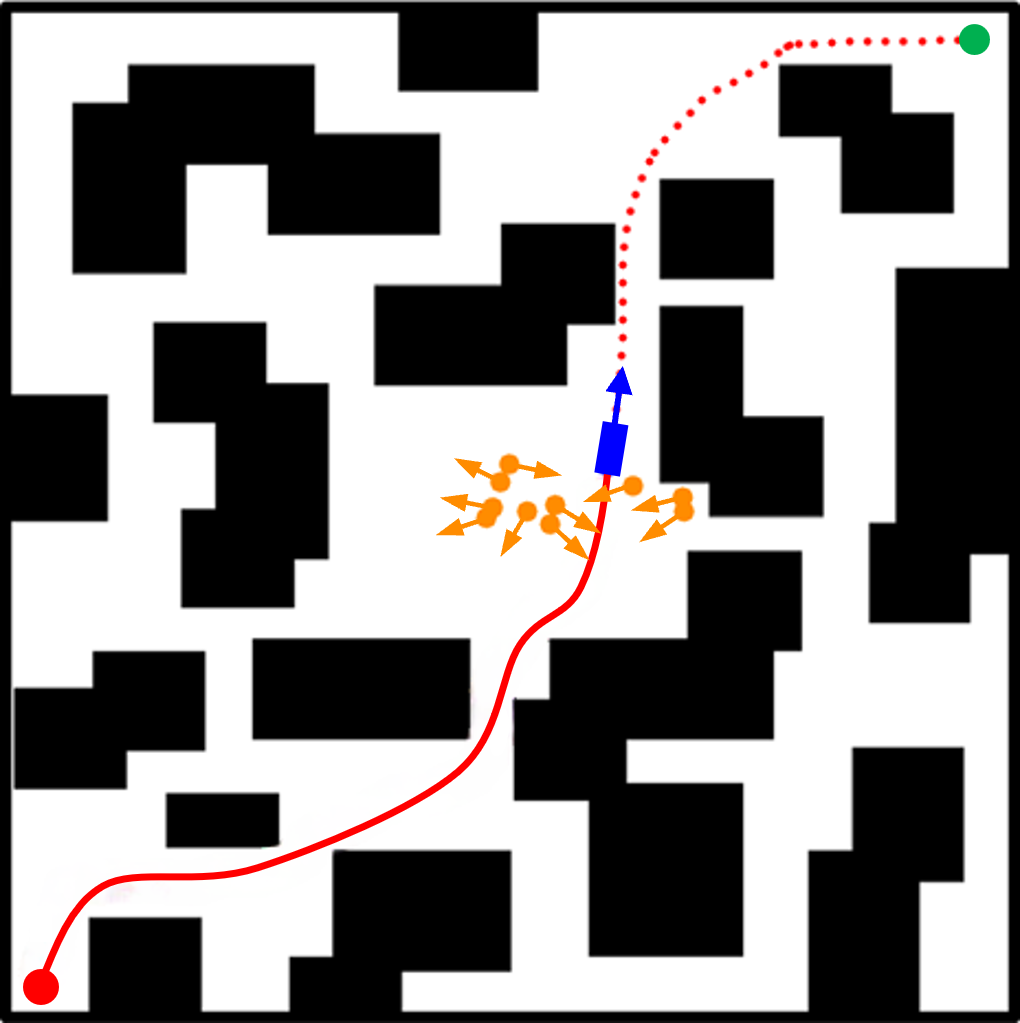}}
    \subfigure[Map \uppercase\expandafter{\romannumeral4}.]{\includegraphics[width=0.49\columnwidth]{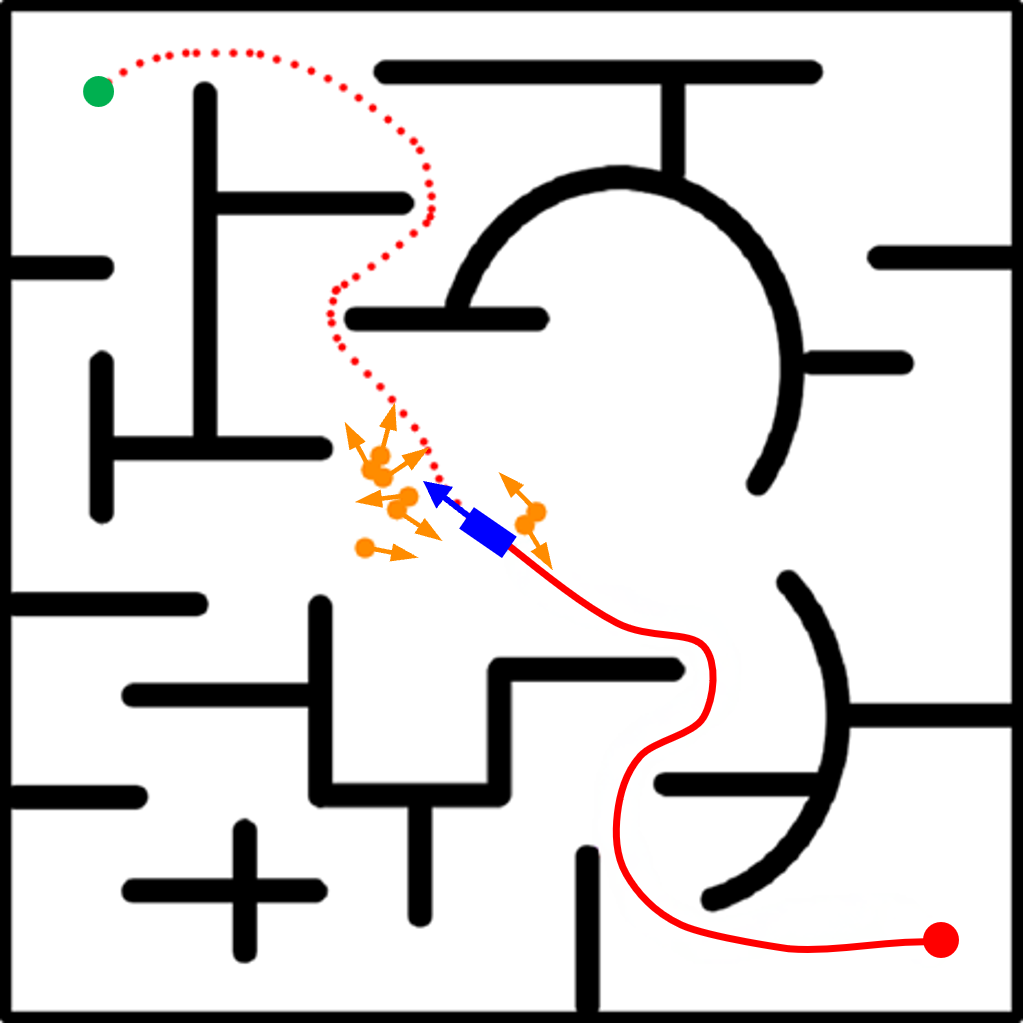}}
\caption{Four different simulation maps. (a) Map \uppercase\expandafter{\romannumeral1}, (b) Map \uppercase\expandafter{\romannumeral2}, (c) Map \uppercase\expandafter{\romannumeral3}, and (d) Map \uppercase\expandafter{\romannumeral4}. In these maps, black squares or lines represent obstacles, while the red and green points represent the start and goal points. The orange points denote dynamic pedestrians, and the blue squares symbolize the robot.}
\label{map}
\end{figure}
In this section, we evaluate the performance of our proposed Multi-Risk-RRT algorithm compared to two popular motion planning algorithms: Risk-RRT and Bi-Risk-RRT.

\subsection{Experiment Setup}
\begin{figure}[htb]
\centering
    \includegraphics[width=1\columnwidth]{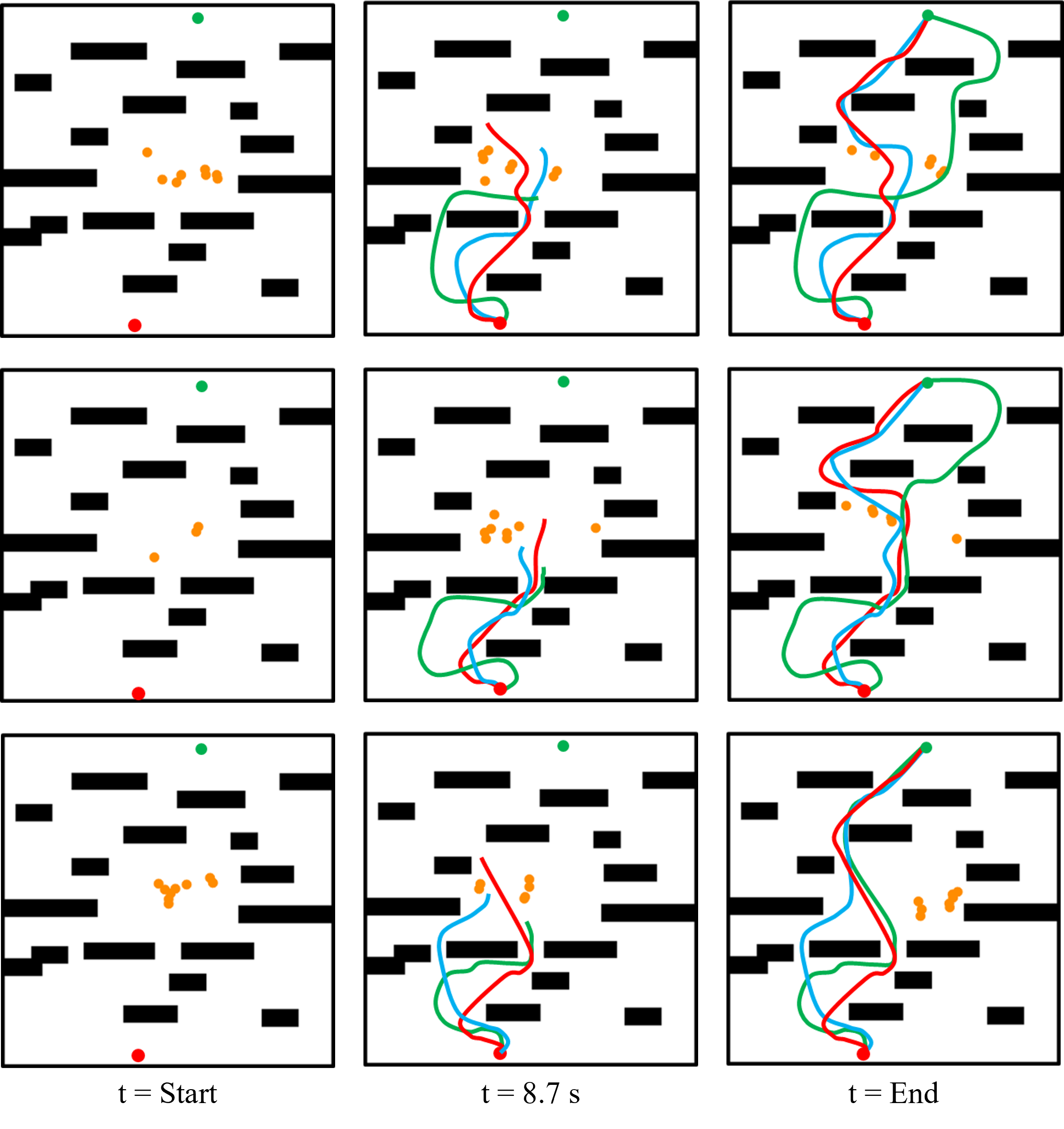}
\caption{Motion planning algorithms in dynamic environments. The red and green dots are markers for the start and goal points, respectively, while the yellow dots signify moving crowds. The trajectories are color-coded to indicate the algorithm used for planning: red represents Multi-Risk-RRT, blue denotes Bi-Risk-RRT, and green means the Risk-RRT algorithm.}
\label{crowds}
\end{figure}

The evaluation is conducted in four designed simulation maps, each reflecting complexities often encountered in airport settings, as illustrated in Fig. \ref{map}. Considering the traits of airports, such as passenger flow in terminals, security checkpoints, and boarding gates, each algorithm is tested on these simulated maps under a static environment and three dynamic crowd scenarios, as depicted in Fig. \ref{crowds}. In the figure, robots use different planning algorithms to navigate through dynamic environments. The rows are categorized by varying crowd conditions: the first row represents Crowds \uppercase\expandafter{\romannumeral1}, the second Crowds \uppercase\expandafter{\romannumeral2}, and the third Crowds \uppercase\expandafter{\romannumeral3}. All these scenarios operate within Map \uppercase\expandafter{\romannumeral1}. These three trajectories for the dynamic crowds are derived from the UCY database \cite{ucy} to simulate the real world, considering situations like peak travel times, boarding and deboarding zones, and emergency evacuations. 

To assess the performance of the algorithms, we employ three metrics: Success Rate, Execution Time, and Trajectory Length. The Success Rate represents the ratio of successful planning attempts to the total number of experiments (considering planning timeout as a failure if it exceeds one hour). Execution Time denotes the time it takes for the robot to traverse from the start state to the goal state. Trajectory Length represents the total length of the trajectory taken by the robot from the start state to the goal state.

The map size for all four environments is 800 $\times$ 800 pixels, with a map resolution of 0.054, corresponding to an actual size of 43.2m $\times$ 43.2m. Tab. \ref{tab:parameter} provides a detailed list of the robot's parameters for simulated and real-world experiments. Here, the term $size$ denotes the spatial dimensions in both the simulation and the real-world environments. The parameters $v_{max}$, $a_{max}$, $\omega_{max}$, and $\alpha_{max}$ correspond to the maximum linear velocity, maximum linear acceleration, maximum angular velocity, and maximum angular acceleration, respectively. Risk-RRT, Bi-Risk-RRT, and Multi-Risk-RRT are implemented in Python using the same planning framework. The experiments are conducted on an AMD EPYC 7413 24-core Processor. Each algorithm is executed 50 times in each scenario to ensure reliable statistical analysis.

\begin{figure}[htb]
\centering
    \includegraphics[width=0.8\columnwidth]{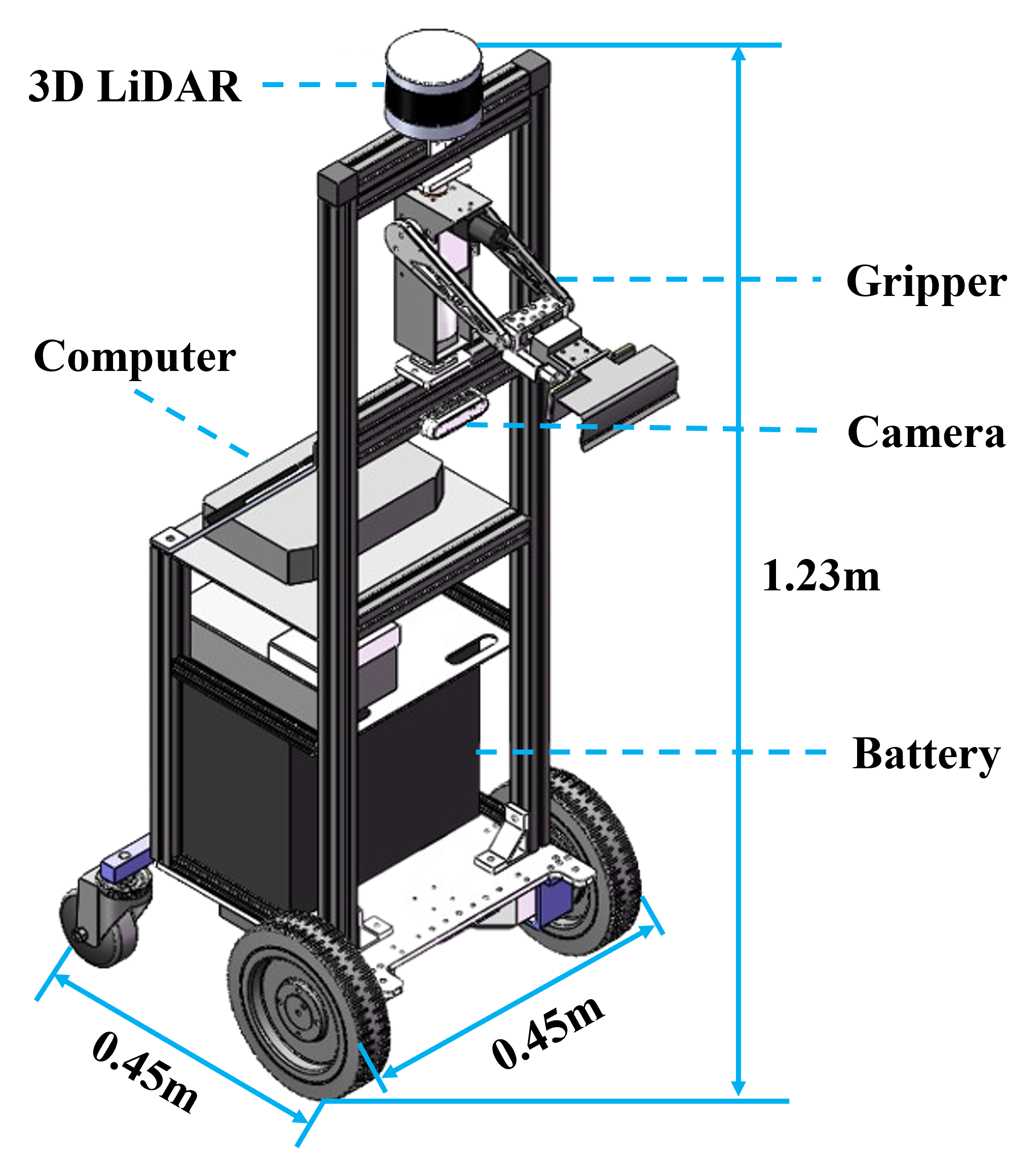}
\caption{Experimental platform for real-world research.}
\label{robot}
\end{figure}

As shown in Fig. \ref{robot}, the experimental platform utilized for real-world research is a luggage trolley collection robot. The robot has dimensions of 0.45m $\times$ 0.45m $\times$ 1.2m and is equipped with various sensors, including a 3D LiDAR and a camera. Controlled by an onboard computer, the robot uses a gripper to collect luggage trolleys.
\begin{table}[htb]
\centering
\caption{Parameters of the Robot Used in Both Simulation and Real-World Experiments}
\label{tab:parameter}
\resizebox{0.48\textwidth}{!}{%
\renewcommand{\arraystretch}{2.5}
\Huge 
\begin{tabular}{cccccc}
\hline
\textbf{}           & \bm{$size (m)$} & \bm{$v_{max} {(m/s)}$} & \bm{$a_{max} (m/s^2)$} & \bm{$\omega_{max} (rad/s)$} & \bm{$\alpha_{max} (rad/s^2)$} \\ \hline
\textbf{Simulaton}  & 43.2 $\times$ 43.2       & 1                & 0.5              & 0.5                & 0.5                \\ \hline
\textbf{Real-world} & 11.5 $\times$ 6.1        & 0.7              & 1                & 0.5                & 0.5                \\ \hline
\end{tabular}
}
\end{table}

\subsection{Experiment Results}
The Success Rate, Execution Time, and Trajectory Length are calculated based on 50 experiments, and the results are presented in Tab. \ref{tab:table1} and Tab. \ref{tab:table2}. It should be noted that since there are failure cases, the actual mean and standard deviation could be higher than the calculated values in the table. However, this does not affect the overall conclusions of the experiments. For example, in Map \uppercase\expandafter{\romannumeral4}, Multi-Risk-RRT achieves a success rate of 100\%, whereas the success rates of Risk-RRT and Bi-Risk-RRT are less than 100\%. However, the mean and standard deviation of Execution Time for Risk-RRT are still lower than Bi-Risk-RRT and significantly lower than those of Multi-Risk-RRT.

Multi-Risk-RRT demonstrates a 100\% Success Rate in all simulation maps in dynamic and static environments. In the four simulated maps, Bi-Risk-RRT outperforms Risk-RRT, but Multi-Risk-RRT exhibits significantly superior performance compared to Bi-Risk-RRT. This result indicates that the efficiency can be improved from single-direction exploration to bi-directional exploration and our enhanced multi-directional exploration further and substantially enhances the exploration efficiency. Regarding Execution Time, Multi-Risk-RRT outperforms Risk-RRT and Bi-Risk-RRT across all four maps under static and dynamic conditions. Furthermore, as indicated in Tab. \ref{tab:table1} and Tab. \ref{tab:table2}, the Trajectory Length achieved by Multi-Risk-RRT is on par with the other two algorithms. Therefore, we improve planning efficiency without compromising the quality of the Trajectory Length.

\begin{table*}[htb]\tiny
\centering
\caption{Evaluating Success Rate, Execution Time, and Trajectory Length: A Comparative Study of Three Algorithms in Static and Crowded \uppercase\expandafter{\romannumeral1} Environments}
\label{tab:table1}
\resizebox{1\textwidth}{!}{%
\begin{tabular}{cccccccc}
\hline
\multirow{2}{*}{\textbf{Map}}        & \multirow{2}{*}{\textbf{Method}} & \multicolumn{3}{c}{\textbf{Static}}                                                    & \multicolumn{3}{c}{\textbf{Crowds \uppercase\expandafter{\romannumeral1}}}                                                  \\ \cline{3-8} 
                                     &                                  & \textbf{Success Rate} & \textbf{Execution Time(s)}     & \textbf{Trajectory Length(m)} & \textbf{Success Rate} & \textbf{Execution Time(s)}     & \textbf{Trajectory Length(m)} \\ \hline
\multirow{3}{*}{\textbf{Map \uppercase\expandafter{\romannumeral1}}}  & \textbf{Risk-RRT}                & 100\%                  & 245.25 ± 28.71                 & 56.72 ± 1.89                  & 100\%                  & 280.23 ± 35.62                 & 57.34 ± 2.64                  \\
                                     & \textbf{Bi-Risk-RRT}             & 100\%                  & 68.10 ± 19.61                  & 62.11 ± 3.29                  & 100\%                  & 107.87 ± 20.65                 & 64.49 ± 3.94                  \\
                                     & \textbf{Multi-Risk-RRT (Ours)}             & 100\%                  & \underline{\textbf{20.50 ± 8.64}}   & 59.87 ± 5.66                  & 100\%                  & \underline{\textbf{37.15 ± 13.91}}   & 59.66 ± 5.18                  \\ \hline
\multirow{3}{*}{\textbf{Map \uppercase\expandafter{\romannumeral2}}} & \textbf{Risk-RRT}                & 78\%                  & 2024.65 ± 167.16               & 62.73 ± 1.75                  & 74\%                  & 2315.91 ± 199.16               & 62.74 ± 1.63                  \\
                                     & \textbf{Bi-Risk-RRT}             & 100\%                  & 100.59 ± 45.91                & 61.25 ± 1.07                  & 96\%                  & 401.78 ± 144.78                & 61.23 ± 1.29                  \\
                                     & \textbf{Multi-Risk-RRT (Ours)}             & 100\%                  & \underline{\textbf{39.45 ± 19.17}}   & 61.44 ± 1.44                  & \underline{\textbf{100\%}}   & \underline{\textbf{126.75 ± 109.35}} & 60.89 ± 1.26                  \\ \hline
\multirow{3}{*}{\textbf{Map \uppercase\expandafter{\romannumeral3}}} & \textbf{Risk-RRT}                & 96\%                  & 1775.32 ± 127.40               & 70.48 ± 2.08                  & 84\%                  & 1891.22 ± 149.82               & 71.16 ± 2.26                  \\
                                     & \textbf{Bi-Risk-RRT}             & 96\%                  & 300.92 ± 94.42                & 69.14 ± 1.84                  & 96\%                  & 488.58 ± 124.24                & 70.06 ± 2.62                  \\
                                     & \textbf{Multi-Risk-RRT (Ours)}             & \underline{\textbf{100\%}}   & \underline{\textbf{158.78 ± 53.87}} & 79.49 ± 7.13                 & \underline{\textbf{100\%}}   & \underline{\textbf{237.65 ± 73.36}} & 83.09 ± 7.24                 \\ \hline
\multirow{3}{*}{\textbf{Map \uppercase\expandafter{\romannumeral4}}} & \textbf{Risk-RRT}                & 68\%                  & 2719.43 ± 106.41               & 76.50 ± 4.32                  & 58\%                  & 2939.00 ± 124.89               & 76.06 ± 3.69                  \\
                                     & \textbf{Bi-Risk-RRT}             & 92\%                  & 1177.60 ± 200.06               & 88.46 ± 2.80                  & 86\%                  & 1488.36 ± 207.81               & 89.28 ± 2.40                  \\
                                     & \textbf{Multi-Risk-RRT (Ours)}             & \underline{\textbf{100\%}}   & \underline{\textbf{156.59 ± 79.73}} & 77.52 ± 3.41                  & \underline{\textbf{100\%}}   & \underline{\textbf{334.71 ± 103.47}} & 78.75 ± 3.64                  \\ \hline
\end{tabular}
}
\end{table*}

\begin{table*}[htb]\tiny
\centering
\caption{Evaluating Success Rate, Execution Time, and Trajectory Length: A Comparative Study of Three Algorithms in Crowded \uppercase\expandafter{\romannumeral2} and Crowded \uppercase\expandafter{\romannumeral3} Environments}
\label{tab:table2}
\resizebox{1\textwidth}{!}{%
\begin{tabular}{cccccccc}
\hline
\multirow{2}{*}{\textbf{Map}}        & \multirow{2}{*}{\textbf{Method}} & \multicolumn{3}{c}{\textbf{Crowds  \uppercase\expandafter{\romannumeral2}}}                                                & \multicolumn{3}{c}{\textbf{Crowds \uppercase\expandafter{\romannumeral3}}}                                                \\ \cline{3-8} 
                                     &                                  & \textbf{Success Rate} & \textbf{Execution Time(s)}     & \textbf{Trajectory Length(m)} & \textbf{Success Rate} & \textbf{Execution Time(s)}     & \textbf{Trajectory Length(m)} \\ \hline
\multirow{3}{*}{\textbf{Map \uppercase\expandafter{\romannumeral1}}}  & \textbf{Risk-RRT}                & 100\%                     & 392.64 ± 51.46                & 59.37 ± 2.59                  & 100\%                     & 365.74 ± 50.13                & 57.57 ± 3.08                  \\
                                     & \textbf{Bi-Risk-RRT}             & 100\%                     & 127.92 ± 31.86                 & 66.08 ± 3.55                  & 100\%                     & 96.24 ± 25.39                  & 62.22 ± 3.48                  \\
                                     & \textbf{Multi-Risk-RRT (Ours)}             & 100\%                     & \underline{\textbf{49.94 ± 23.29}}   & 60.57 ± 5.71                  & 100\%                     & \underline{\textbf \textbf{24.53 ± 9.45}}   & 58.42 ± 4.80                  \\ \hline
\multirow{3}{*}{\textbf{Map \uppercase\expandafter{\romannumeral2}}} & \textbf{Risk-RRT}                & 64\%                  & 2028.28 ± 177.14               & 61.90 ± 1.15                  & 68\%                  & 2298.81 ± 195.85               & 62.59 ± 1.44                  \\
                                     & \textbf{Bi-Risk-RRT}             & 100\%                     & 179.76 ± 66.40                & 61.21 ± 1.09                  & 96\%                  & 306.96 ± 124.01                & 61.91 ± 1.30                  \\
                                     & \textbf{Multi-Risk-RRT (Ours)}             & 100\%                    & \underline{\textbf{57.25 ± 25.77}}   & 60.84 ± 1.21                  & \underline{\textbf {100\%}}      & \underline{ \textbf{73.73 ± 52.26}}  & 61.92 ± 1.55                  \\ \hline
\multirow{3}{*}{\textbf{Map \uppercase\expandafter{\romannumeral3}}} & \textbf{Risk-RRT}                & 76\%                  & 1883.76 ± 175.01               & 70.10 ± 1.97                  & 90\%                   & 2028.43 ± 178.49               & 73.04 ± 2.94                  \\
                                     & \textbf{Bi-Risk-RRT}             & 84\%                  & 874.21 ± 209.38                & 72.89 ± 4.16                  & 94\%                  & 599.17 ± 174.38                & 69.05 ± 1.77                  \\
                                     & \textbf{Multi-Risk-RRT (Ours)}             & \underline{\textbf{100\%}}      & \underline{\textbf{246.81 ± 67.11}} & 81.02 ± 7.36                 & \underline{\textbf{100\%}}      & \underline{\textbf{202.81 ± 45.53}} & 79.92 ± 6.24                  \\ \hline
\multirow{3}{*}{\textbf{Map \uppercase\expandafter{\romannumeral4}}} & \textbf{Risk-RRT}                & 46\%                  & 3010.59 ± 107.85               & 82.15 ± 5.09                  & 60\%                   & 2812.11 ± 144.51               & 76.13 ± 3.38                  \\
                                     & \textbf{Bi-Risk-RRT}             & 78\%                  & 1259.68 ± 199.17               & 90.34 ± 2.69                  & 82\%                  & 1364.15 ± 220.13               & 90.74 ± 2.06                  \\
                                     & \textbf{Multi-Risk-RRT (Ours)}             & \underline{\textbf{100\%}}      & \underline{\textbf{290.28 ± 98.52}} & 80.11 ± 4.13                  & \underline{\textbf{100\%}}      & \underline{\textbf{312.77 ± 115.52}} & 78.73 ± 3.95                  \\ \hline
\end{tabular}
}
\end{table*}

In addition to simulations, we also carried out a real-world experiment, illustrated in Fig. \ref{real-world}. For a comprehensive view of the experimental process, please refer to the video in the supplementary materials. This experiment featured three static obstacles and five pedestrians. Consistent with our simulation findings, the experimental results show that Multi-Risk-RRT achieved the lowest Execution Time among the tested algorithms.
\begin{figure*}[htb]
\centering
    \includegraphics[width=2\columnwidth]{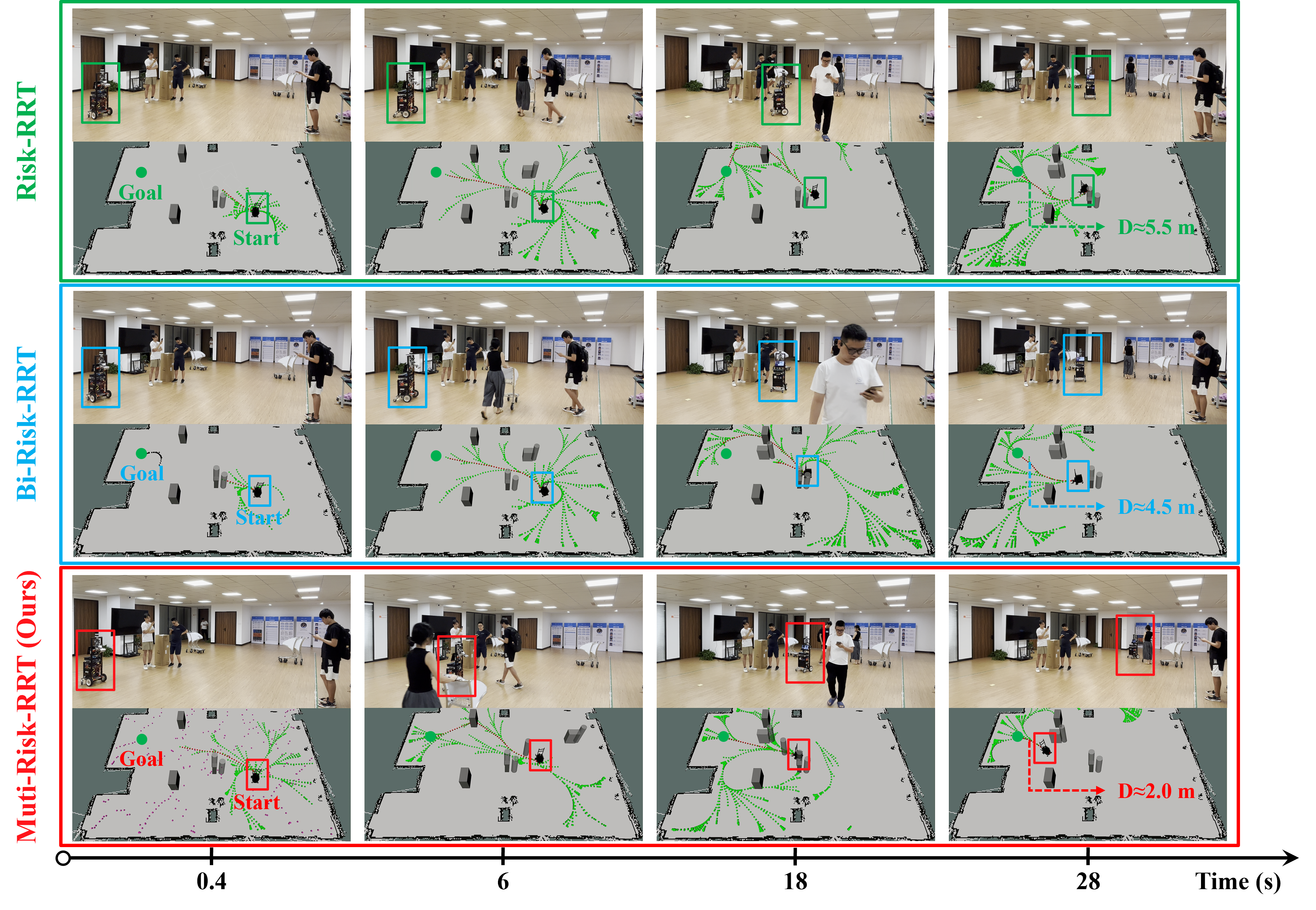}
\caption{Screenshots capturing key moments from the real-world experimental process. Images have been captured at four distinct time points—0.4 s, 6 s, 18 s, and 28 s—to overview the entire planning process. The large boxes in three distinct colors signify the outcomes of three different algorithms. Correspondingly, the smaller boxes in the same color schemes represent the robot's position at various time intervals under each algorithm. The robot's initial position in the first column is the start point, while the dot symbolizes the goal point. The D in the last column indicates the robot's approximate distance from the goal.}
\label{real-world}
\end{figure*}

\section{Discussion}
In this section, we will delve into a detailed analysis of the factors contributing to the performance enhancement of Multi-Risk-RRT. These factors can be attributed to two key aspects. Firstly, the MSHS strategy improves the algorithm's capacity to acquire comprehensive global information. Secondly, the heuristic information rapid updating is crucial in improving the algorithm's efficiency and robustness.

\subsection{MSHS Strategy}
\begin{figure*}[htb]
\centering
    \includegraphics[width=2\columnwidth]{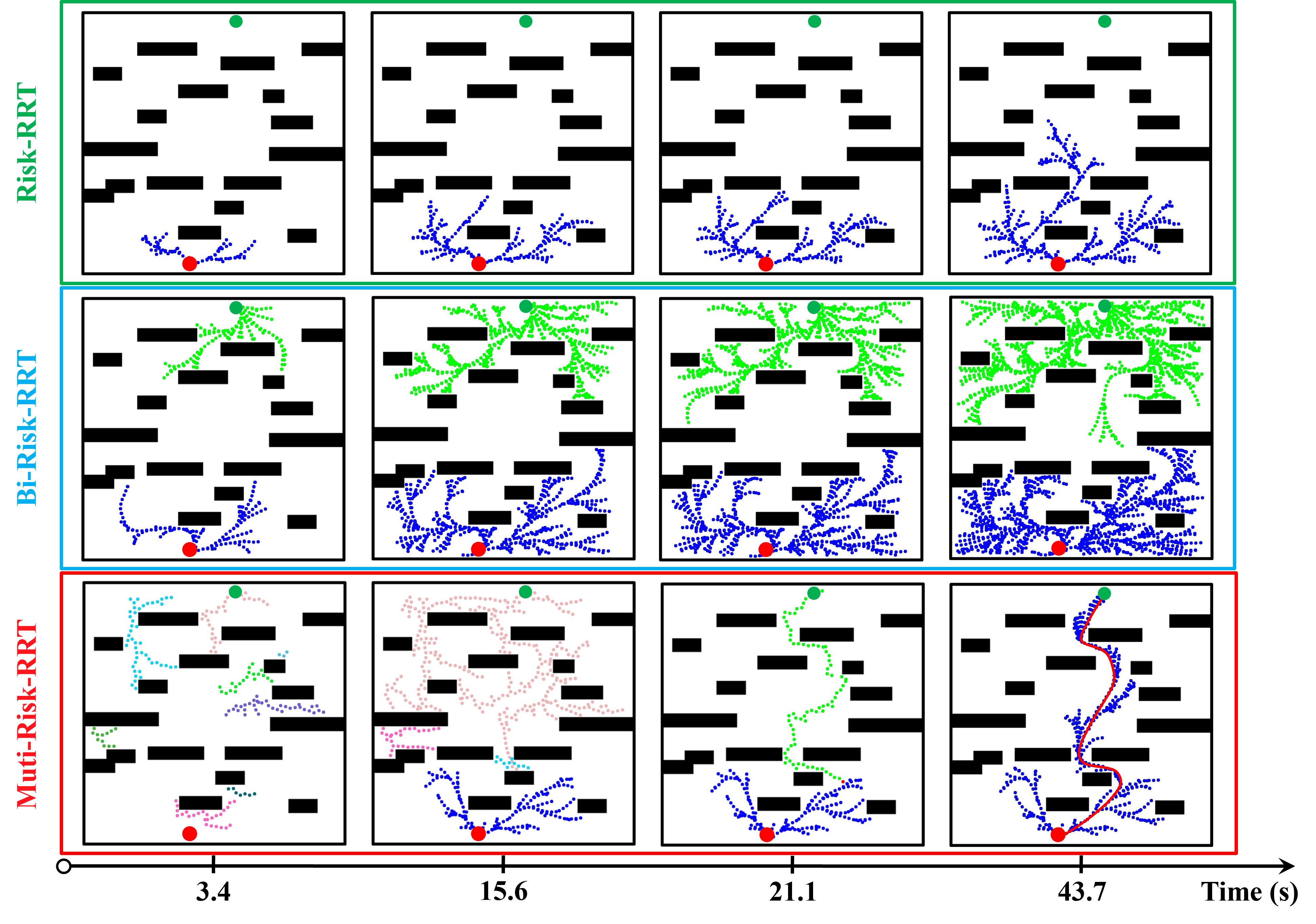}
\caption{Comparison of the search processes of three algorithms. The first row illustrates the performance of Risk-RRT, the second row represents Bi-Risk-RRT, and the third row shows Multi-Risk-RRT. The algorithms are evaluated at four distinct time points: t = 3.4 s, t = 15.6 s, t = 21.1 s, and t = 43.7 s.}
\label{compare}
\end{figure*}

The MSHS strategy enhances the algorithm's global exploration ability. Risk-RRT, as a single-tree approach, concentrates its exploration near the start point, resulting in limited spatial coverage within a given period. Bi-Risk-RRT improves on this by employing dual-tree growth, allowing exploration near the start and end points. This bi-directional search strategy enhances the exploration capability compared to Risk-RRT. With its incremental multi-tree growth, Multi-Risk-RRT takes exploration to a higher level by expanding its coverage to nearly every position in the space, as shown in Fig. \ref{compare}. The figure captures snapshots of the search process for three algorithms at various time points. Multi-Risk-RRT, employing the MSHS strategy, achieves a broader search scope and advances more aggressively toward the target. Although Bi-Risk-RRT improves its search efficiency through bi-directional growth, it falls notably short of Multi-Risk-RRT's performance. By MSHS strategy, Multi-Risk-RRT can quickly acquire different parts of the space from the sub-tree, enabling quick adaptation and environmental knowledge updates. This strategy significantly improves the efficiency and robustness of the algorithm compared to Bi-Risk-RRT and Risk-RRT. In summary, Multi-Risk-RRT's ability to rapidly acquire heuristic information allows for quicker identification of an initial feasible trajectory compared to both Bi-Risk-RRT and Risk-RRT.

\subsection{Heuristic Information Updating}
\begin{figure}[htb]
\centering
    \subfigure[Static.]{\includegraphics[width=0.49\columnwidth]{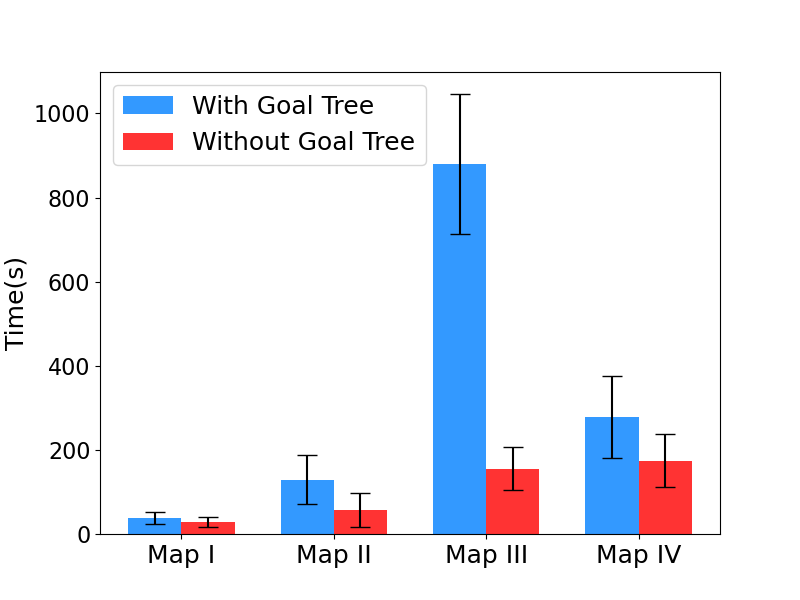}}
    \subfigure[Crows \uppercase\expandafter{\romannumeral1}.]{\includegraphics[width=0.49\columnwidth]{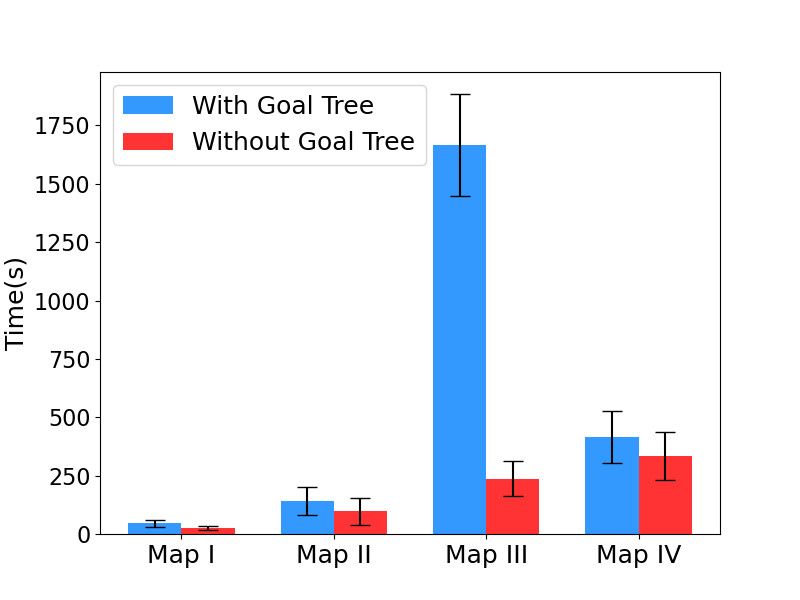}}
    \subfigure[Crows \uppercase\expandafter{\romannumeral2}.]{\includegraphics[width=0.49\columnwidth]{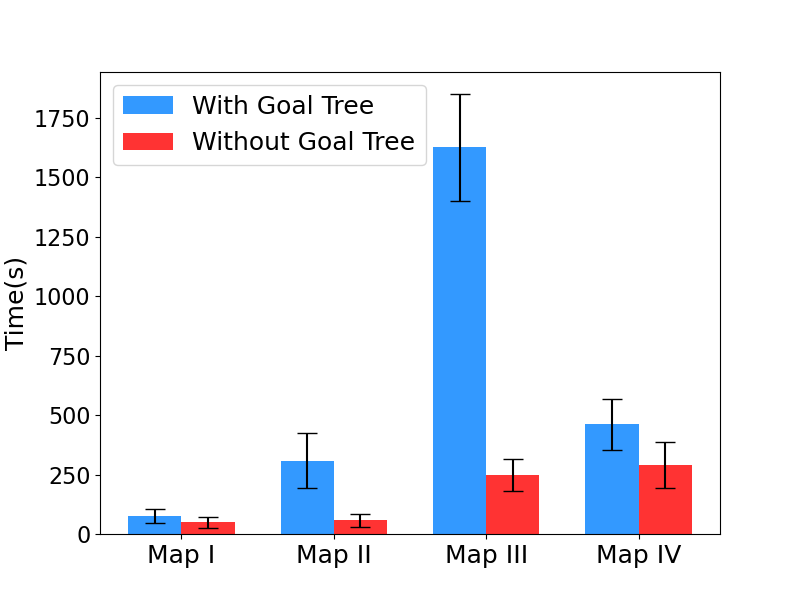}}
    \subfigure[Crows \uppercase\expandafter{\romannumeral3}.]{\includegraphics[width=0.49\columnwidth]{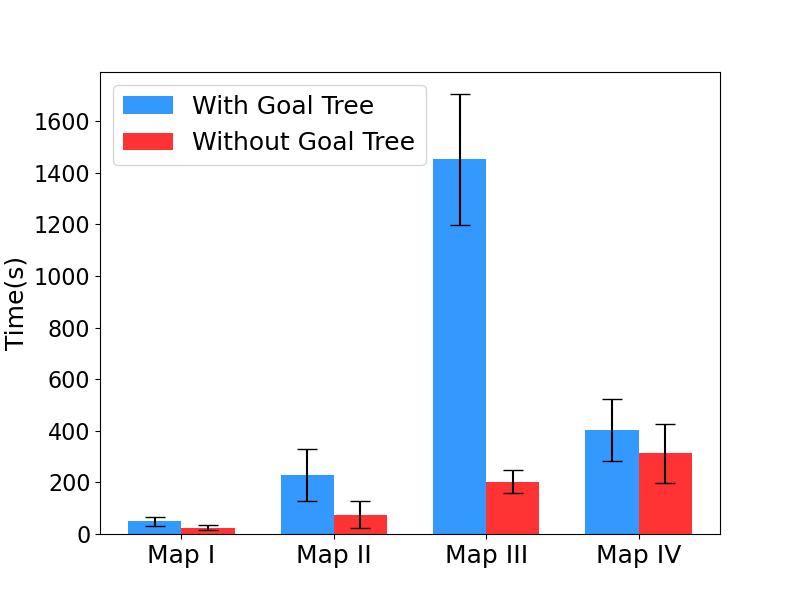}}
\caption{Experimental results comparing the Execution Time of Multi-Risk-RRT with and without goal tree in different scenarios. (a) Static, (b) Crows \uppercase\expandafter{\romannumeral1}, (c) Crows \uppercase\expandafter{\romannumeral2}, (d) Crows \uppercase\expandafter{\romannumeral3}.}
\label{with or without goal tree}
\end{figure}

One significant improvement of Multi-Risk-RRT is its utilization of heuristic information from the direction of the goal. When such heuristic information is available, the algorithm guides the robot more greedily toward the target direction. However, it is essential to note that there is no guarantee that a feasible trajectory can be found solely based on the heuristic information from the goal direction. To investigate the impact of this heuristic information, we experiment by comparing the Execution Time of Multi-Risk-RRT with the condition that the goal tree is retained or deleted. Retention involves using the heuristic information multiple times, while deletion means generating new heuristic information from the state space. The results, shown in Fig. \ref{with or without goal tree}, present the mean and standard deviation of the sixteen scenarios. The mean values indicate that updating the heuristic information has a lower average Execution Time than retaining it. Additionally, the trend of the standard deviation is similar to the mean value, indicating that retaining the goal tree offers greater robustness compared to its deletion.

Interestingly, preserving the goal tree yields particularly poor performance on Map \uppercase\expandafter{\romannumeral3} compared to other simulated maps. Map \uppercase\expandafter{\romannumeral3} is a complex environment with numerous inflection points that can hinder the robot's utilization of heuristic information. The initial heuristic information may not effectively guide the robot due to uneven sampling or a lack of exploration in other areas. Consequently, the robot may become trapped by invalid heuristic information. In such environments, iterative updates of the heuristic information play a critical role. Even if the initial heuristic information is ineffective, the iterative updates ensure continuous state space exploration, effectively guiding the robot toward the target direction.

\section{Conclusion and Future Work}
This article presents the MSHS strategy that combines Multi-directional Searching with Heuristic Sampling. This integrated strategy efficiently merges heuristic information from dynamic sub-trees into the rooted tree, bypassing TBVP solvers' limitations. During Multi-directional Searching, the state obtained from the uniform sample is added to either the rooted tree or any sub-tree. In Heuristic Sampling, the rooted tree gradually collects information from sub-trees, facilitating its growth toward the goal state. Furthermore, Heuristic Information Updating equips Multi-Risk-RRT with the ability to adapt to complex and dynamic environments effectively. We evaluate the proposed algorithm across simulations and real-world environmental studies. The outcomes underline the capacity of the Multi-Risk-RRT algorithm to enhance motion planning performance in both static and dynamic environments. Moreover, we deliberate on the MSHS strategy and the impact of heuristic information updates, demonstrating the practical applicability and robustness of the proposed algorithm.

In future work, we plan to deploy the Multi-Risk-RRT algorithm at Hong Kong International Airport and Shenzhen Baoan International Airport. We aim further to validate the algorithm's effectiveness and robustness in real-world airport conditions.

\bibliographystyle{IEEEtran} 

\bibliography{refs} 

\begin{thebibliography}{10}
\providecommand{\url}[1]{#1}
\csname url@samestyle\endcsname
\providecommand{\newblock}{\relax}
\providecommand{\bibinfo}[2]{#2}
\providecommand{\BIBentrySTDinterwordspacing}{\spaceskip=0pt\relax}
\providecommand{\BIBentryALTinterwordstretchfactor}{4}
\providecommand{\BIBentryALTinterwordspacing}{\spaceskip=\fontdimen2\font plus
\BIBentryALTinterwordstretchfactor\fontdimen3\font minus \fontdimen4\font\relax}
\providecommand{\BIBforeignlanguage}[2]{{%
\expandafter\ifx\csname l@#1\endcsname\relax
\typeout{** WARNING: IEEEtran.bst: No hyphenation pattern has been}%
\typeout{** loaded for the language `#1'. Using the pattern for}%
\typeout{** the default language instead.}%
\else
\language=\csname l@#1\endcsname
\fi
#2}}
\providecommand{\BIBdecl}{\relax}
\BIBdecl

\bibitem{real-time}
J.~Wang and M.~Q.-H. Meng, ``Real-time decision making and path planning for robotic autonomous luggage trolley collection at airports,'' \emph{IEEE Transactions on Systems, Man, and Cybernetics: Systems}, vol.~52, no.~4, pp. 2174--2183, 2022.

\bibitem{xiao-icra}
A.~Xiao, H.~Luan, Z.~Zhao, Y.~Hong, J.~Zhao, W.~Chen, J.~Wang, and M.~Q.-H. Meng, ``Robotic autonomous trolley collection with progressive perception and nonlinear model predictive control,'' in \emph{2022 International Conference on Robotics and Automation (ICRA)}, 2022, pp. 4480--4486.

\bibitem{hr}
V.~S. Chirala, K.~Sundar, S.~Venkatachalam, J.~M. Smereka, and S.~Kassoumeh, ``Heuristics for multi-vehicle routing problem considering human-robot interactions,'' \emph{IEEE Transactions on Intelligent Vehicles}, vol.~8, no.~5, pp. 3228--3238, 2023.

\bibitem{apf}
O.~Khatib, ``Real-time obstacle avoidance for manipulators and mobile robots,'' \emph{The international journal of robotics research}, vol.~5, no.~1, pp. 90--98, 1986.

\bibitem{A*}
P.~E. Hart, N.~J. Nilsson, and B.~Raphael, ``A formal basis for the heuristic determination of minimum cost paths,'' \emph{IEEE Transactions on Systems Science and Cybernetics}, vol.~4, pp. 100--107, 1968.

\bibitem{Dijkstra}
E.~W. Dijkstra, ``A note on two problems in connexion with graphs,'' \emph{Numerische Mathematik}, vol.~1, pp. 269--271, 1959.

\bibitem{computation}
J.~H. Reif, ``Complexity of the mover's problem and generalizations,'' \emph{foundations of computer science}, 1979.

\bibitem{RRT}
S.~M. LaValle, ``Rapidly-exploring random trees : a new tool for path planning,'' \emph{The annual research report}, 1998.

\bibitem{PRM}
L.~E. Kavraki, P.~Svestka, J.-C. Latombe, and M.~H. Overmars, ``Probabilistic roadmaps for path planning in high-dimensional configuration spaces,'' in \emph{International Conference on Robotics and Automation}, 1996.

\bibitem{Informed_rrt}
J.~D. Gammell, S.~S. Srinivasa, and T.~D. Barfoot, ``Informed {RRT*}: {Optimal} sampling-based path planning focused via direct sampling of an admissible ellipsoidal heuristic,'' in \emph{Intelligent Robots and Systems}, 2014.

\bibitem{rrt-connect}
J.~J. Kuffner and S.~M. LaValle, ``{RRT-Connect}: {An} efficient approach to single-query path planning,'' in \emph{International Conference on Robotics and Automation}, 2000.

\bibitem{rrdt}
T.~Lai, F.~Ramos, and G.~Francis, ``Balancing global exploration and local-connectivity exploitation with rapidly-exploring random disjointed-trees,'' in \emph{2019 International Conference on Robotics and Automation (ICRA)}.\hskip 1em plus 0.5em minus 0.4em\relax IEEE, 2019, pp. 5537--5543.

\bibitem{tbvp}
H.~B. Keller, \emph{Numerical methods for two-point boundary-value problems}.\hskip 1em plus 0.5em minus 0.4em\relax Courier Dover Publications, 2018.

\bibitem{constrain}
J.-P. Laumond, P.~E. Jacobs, M.~Taix, and R.~M. Murray, ``A motion planner for nonholonomic mobile robots,'' \emph{IEEE Transactions on robotics and automation}, vol.~10, no.~5, pp. 577--593, 1994.

\bibitem{risk-rrt}
C.~Fulgenzi, A.~Spalanzani, C.~Laugier, and C.~Tay, ``Risk based motion planning and navigation in uncertain dynamic environment,'' 2010.

\bibitem{lm-rrt}
W.~Wang, L.~Zuo, and X.~Xu, ``A learning-based {Multi-RRT} approach for robot path planning in narrow passages,'' \emph{Journal of Intelligent and Robotic Systems}, 2018.

\bibitem{b2u}
J.~Wang, W.~Chi, C.~Li, and M.~Q.-H. Meng, ``Efficient robot motion planning using bidirectional-unidirectional {RRT} extend function,'' \emph{IEEE Transactions on Automation Science and Engineering}, 2021.

\bibitem{mt-rrt}
Z.~Sun, J.~Wang, and M.~Q.-H. Meng, ``Multi-tree guided efficient robot motion planning,'' \emph{Procedia Computer Science}, vol. 209, pp. 31--39, 2022.

\bibitem{cave}
B.~Ichter, J.~Harrison, and M.~Pavone, ``Learning sampling distributions for robot motion planning,'' in \emph{2018 IEEE International Conference on Robotics and Automation (ICRA)}.\hskip 1em plus 0.5em minus 0.4em\relax IEEE, 2018, pp. 7087--7094.

\bibitem{gan}
T.~Zhang, J.~Wang, and M.~Q.-H. Meng, ``Generative adversarial network based heuristics for sampling-based path planning,'' \emph{IEEE/CAA Journal of Automatica Sinica}, vol.~9, no.~1, pp. 64--74, 2022.

\bibitem{re-rrt}
K.~Naderi, J.~Rajam{\"a}ki, and P.~H{\"a}m{\"a}l{\"a}inen, ``Rt-rrt* a real-time path planning algorithm based on rrt,'' in \emph{Proceedings of the 8th ACM SIGGRAPH Conference on Motion in Games}, 2015, pp. 113--118.

\bibitem{rrtx}
M.~Otte and E.~Frazzoli, ``Rrtx: Asymptotically optimal single-query sampling-based motion planning with quick replanning,'' \emph{The International Journal of Robotics Research}, vol.~35, no.~7, pp. 797--822, 2016.

\bibitem{mprrt}
M.~Zucker, J.~Kuffner, and M.~Branicky, ``Multipartite rrts for rapid replanning in dynamic environments,'' in \emph{Proceedings 2007 IEEE International Conference on Robotics and Automation}.\hskip 1em plus 0.5em minus 0.4em\relax IEEE, 2007, pp. 1603--1609.

\bibitem{d*}
A.~Stentz, ``Optimal and efficient path planning for partially-known environments,'' in \emph{Proceedings of the 1994 IEEE international conference on robotics and automation}.\hskip 1em plus 0.5em minus 0.4em\relax IEEE, 1994, pp. 3310--3317.

\bibitem{d*-lite}
S.~Koenig and M.~Likhachev, ``D* lite,'' in \emph{Eighteenth national conference on Artificial intelligence}, 2002, pp. 476--483.

\bibitem{nr-rrt}
F.~Meng, L.~Chen, H.~Ma, J.~Wang, and M.~Q.-H. Meng, ``Nr-rrt: Neural risk-aware near-optimal path planning in uncertain nonconvex environments,'' \emph{IEEE Transactions on Automation Science and Engineering}, pp. 1--12, 2022.

\bibitem{risk-dtrrt}
W.~Chi, C.~Wang, J.~Wang, and M.~Q.-H. Meng, ``Risk-dtrrt-based optimal motion planning algorithm for mobile robots,'' \emph{IEEE Transactions on Automation Science and Engineering}, vol.~16, no.~3, pp. 1271--1288, 2019.

\bibitem{bi-risk-rrt}
H.~Ma, F.~Meng, C.~Ye, J.~Wang, and M.~Q.-H. Meng, ``Bi-risk-rrt based efficient motion planning for autonomous ground vehicles,'' \emph{IEEE Transactions on Intelligent Vehicles}, vol.~7, no.~3, pp. 722--733, 2022.

\bibitem{motioncontrol}
Y.~Gan, B.~Zhang, C.~Ke, X.~Zhu, W.~He, and T.~Ihara, ``Research on robot motion planning based on rrt algorithm with nonholonomic constraints,'' \emph{Neural Processing Letters}, vol.~53, no.~4, pp. 3011--3029, 2021.

\bibitem{gmm}
D.~A. Reynolds \emph{et~al.}, ``Gaussian mixture models.'' \emph{Encyclopedia of biometrics}, vol. 741, no. 659-663, 2009.

\bibitem{ucy}
A.~Lerner, Y.~Chrysanthou, and D.~Lischinski, ``Crowds by example,'' in \emph{Computer graphics forum}, vol.~26, no.~3.\hskip 1em plus 0.5em minus 0.4em\relax Wiley Online Library, 2007, pp. 655--664.

\end{thebibliography}

\begin{IEEEbiography}
[{\includegraphics[width=1in,height=1.25in,clip,keepaspectratio]{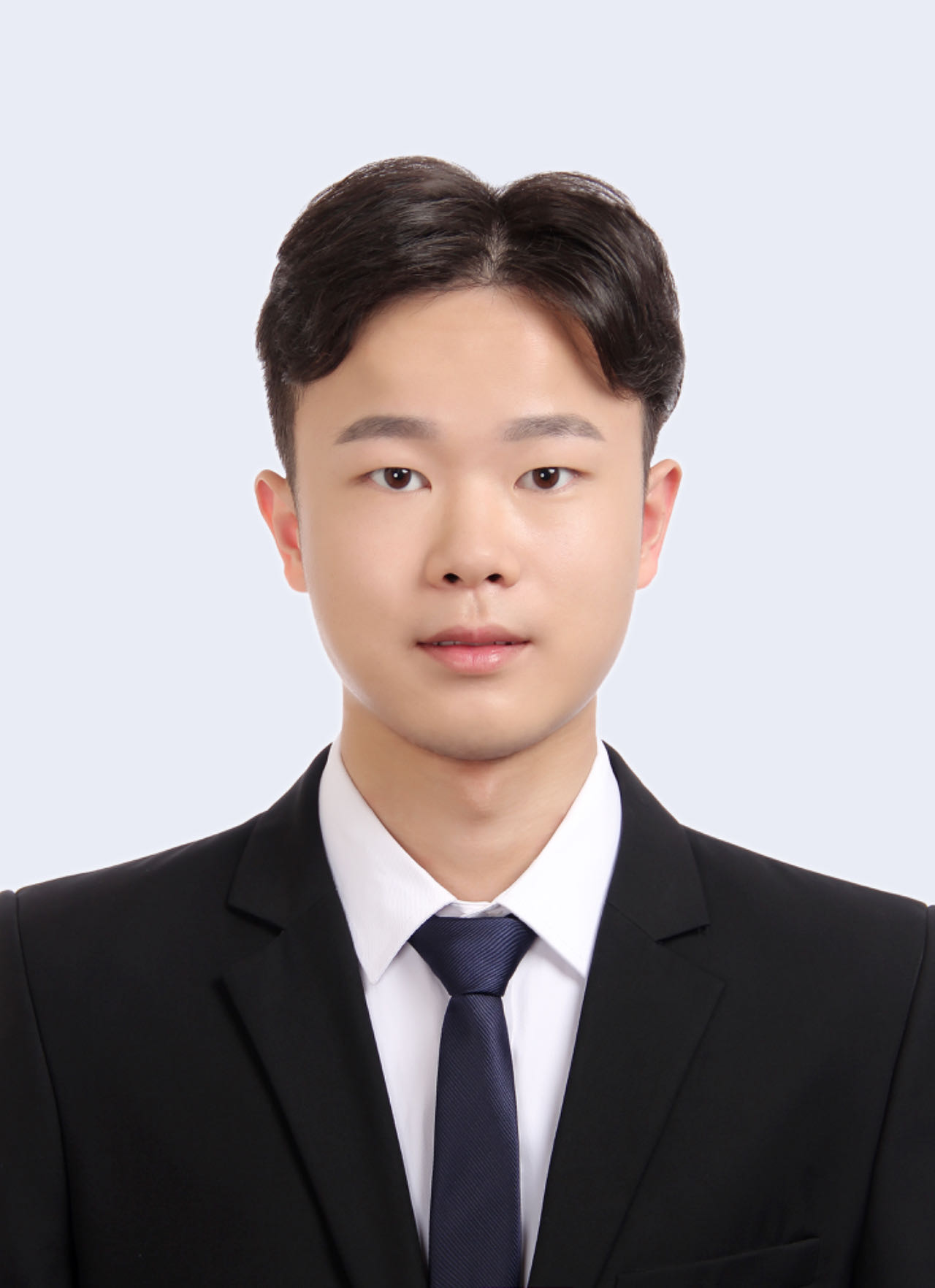}}] 
{Zhirui Sun} received the B.E. degree in information engineering from the Department of Electronic and Electrical Engineering, Southern University of Science and Technology, Shenzhen, China, in 2019. He is currently pursuing the Ph.D. degree with the Department of Electronic and Electrical Engineering, Southern University of Science and Technology, Shenzhen, China. His research interests include robot perception and motion planning.
\end{IEEEbiography}
\vspace{-10 mm}

\begin{IEEEbiography}
[{\includegraphics[width=1in,height=1.25in,clip,keepaspectratio]{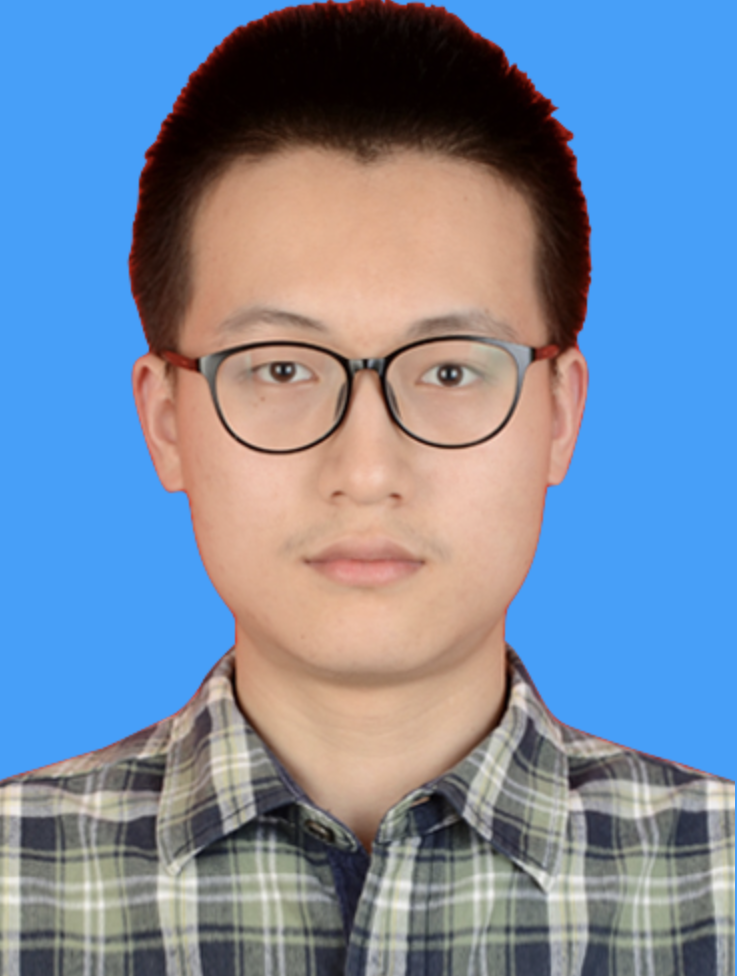}}] 
{Boshu Lei} received the B.E. degree in automation from the Department of Electronic and Information Engineering, Xi'an Jiao Tong University, Xi'an, China, in 2022. He is currently pursuing the M.S. degree with the Department of Engineering and Applied Science, University of Pennsylvania, Philadelphia, U.S.A. His research interests include robots and computer vision.
\end{IEEEbiography}
\vspace{-10 mm}

\begin{IEEEbiography}
[{\includegraphics[width=1in,height=1.25in,clip,keepaspectratio]{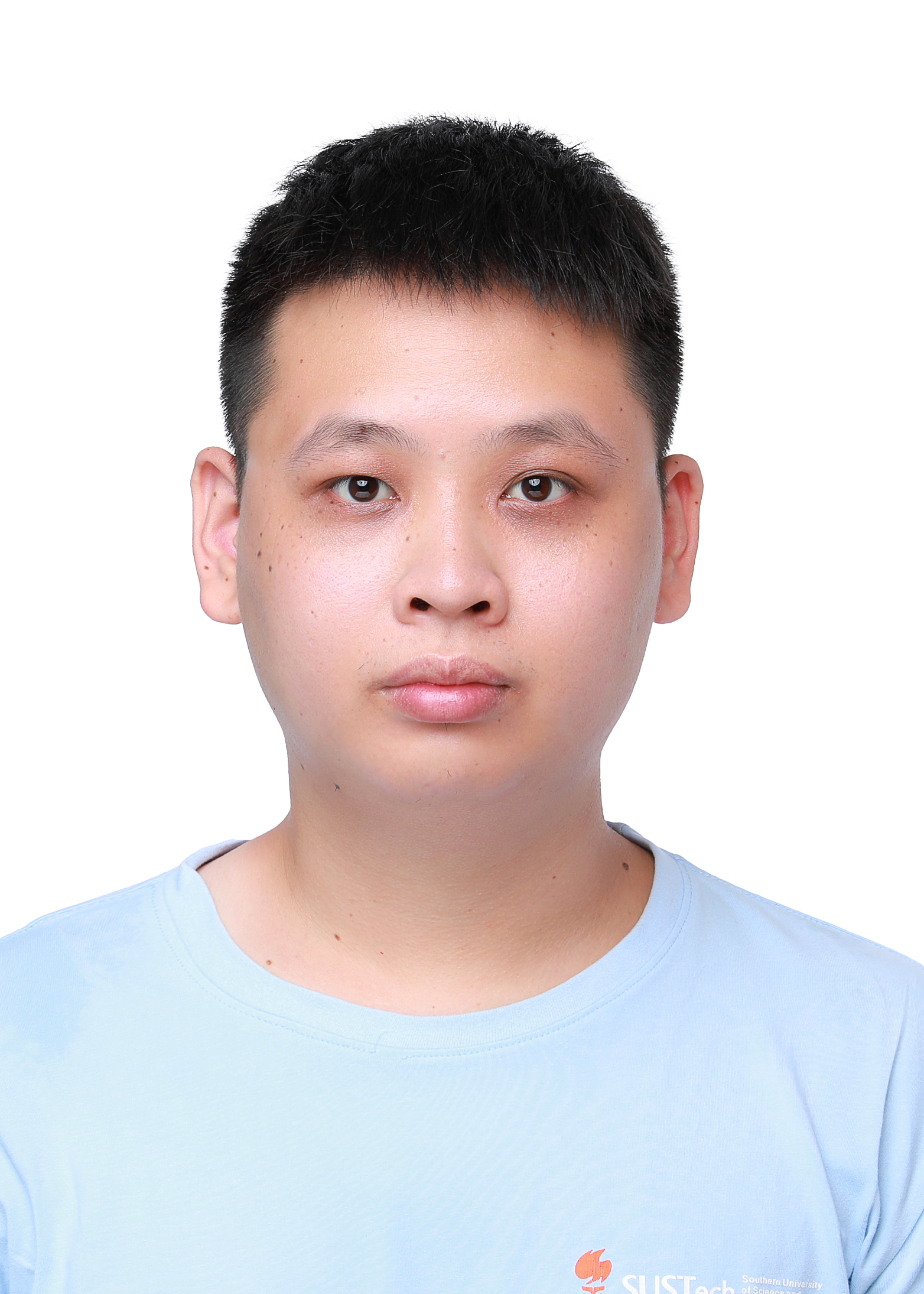}}] 
{Peijia Xie} received the B.E. degree in Electronic Information Engineering from the School of Physics and Telecommunications Engineering, South China Normal University, Guangzhou, China, in 2022. He is currently pursuing the M.S. degree with the Department of Electronic and Electrical Engineering, Southern University of Science and Technology, Shenzhen, China. His primary research interest is autonomous driving.
\end{IEEEbiography}
\vspace{-10 mm}

\begin{IEEEbiography}
[{\includegraphics[width=1in,height=1.25in,clip,keepaspectratio]{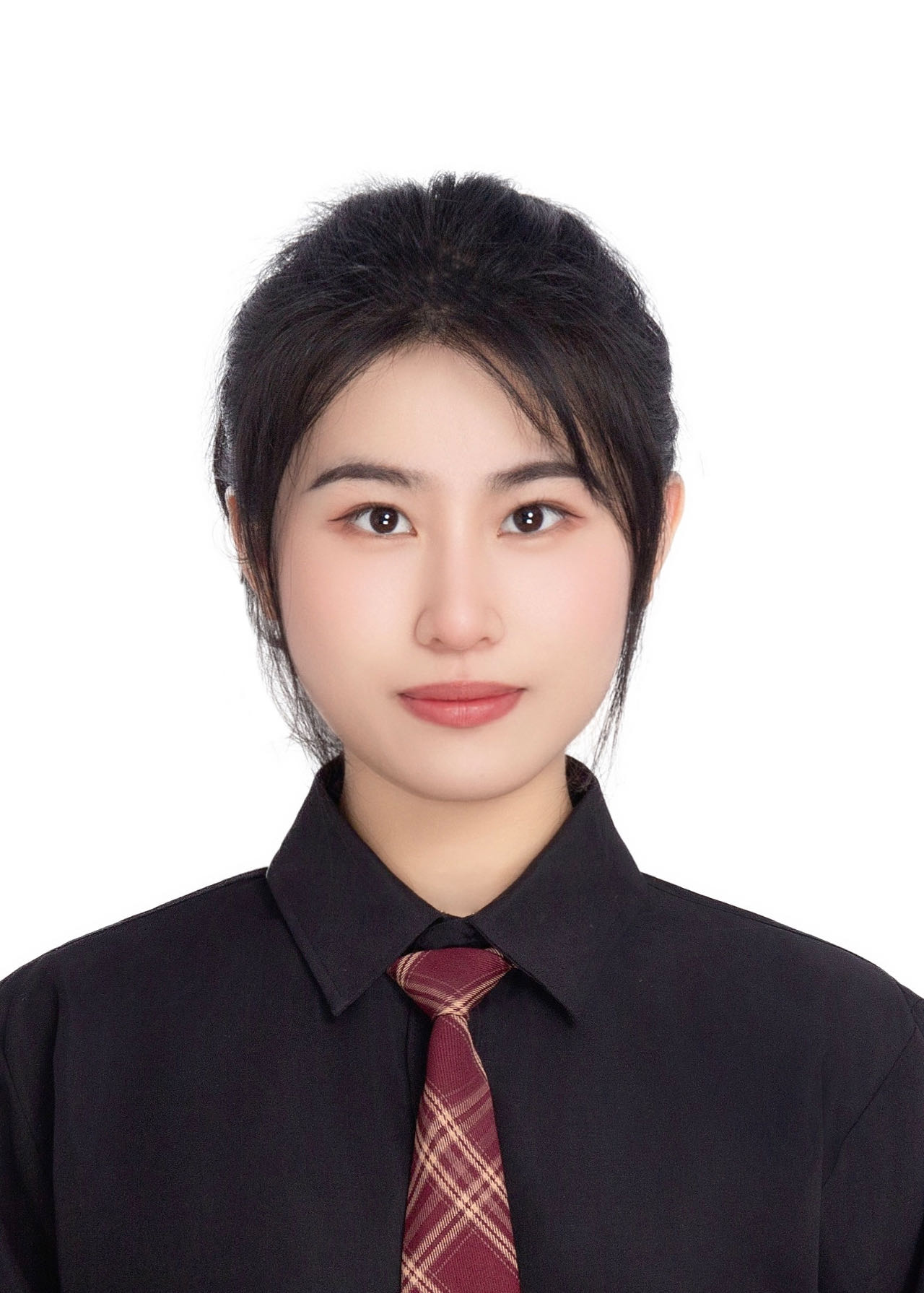}}] 
{Fugang Liu} received the B.E. degree in the Department of Biomedical Engineering, Southern University of Science and Technology, Shenzhen, China, in 2019. She is currently pursuing the Ph.D. degree with the Department of Biomedical Engineering, Shanghai Jiao Tong University, Shanghai, China. Her primary research interest is detecting cellular metabolism using SERS.
\end{IEEEbiography}
\vspace{-10 mm}

\begin{IEEEbiography}
[{\includegraphics[width=1in,height=1.25in,clip,keepaspectratio]{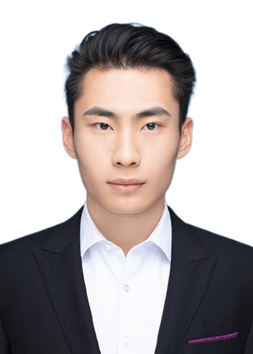}}] 
{Junjie Gao} received the B.E. degree and M.S. degree in Control Science and Engineering from the School of Astronautics, Harbin Institute of Technology, Harbin, China, in 2021 and 2023, respectively. He is currently a research assistant with the Department of Electronic and Electrical Engineering, Southern University of Science and Technology, Shenzhen, China. His research interests include mobile robots, motion planning, and autonomous exploration.
\end{IEEEbiography}
\vspace{-10 mm}

\begin{IEEEbiography}
[{\includegraphics[width=1in,height=1.25in,clip,keepaspectratio]{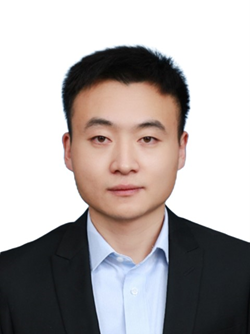}}] 
{Ying Zhang} (Member, IEEE) received the B.S. degree in Automatic Control from Heze University, Heze, China, in 2014, the M.S. degree in Control Theory and Control Engineering from Shandong Jianzhu University, Jinan, China, in 2017, and the Ph.D. degree in Control Theory and Control Engineering from Shandong University, Jinan, China, in 2021. 

He is currently a Lecturer with the School of Electrical Engineering, Yanshan University. His current research interests include intelligent robot system, visual perception, and environment modeling.
\end{IEEEbiography}
\vspace{-10 mm}

\begin{IEEEbiography}
[{\includegraphics[width=1in,height=1.25in,clip,keepaspectratio]{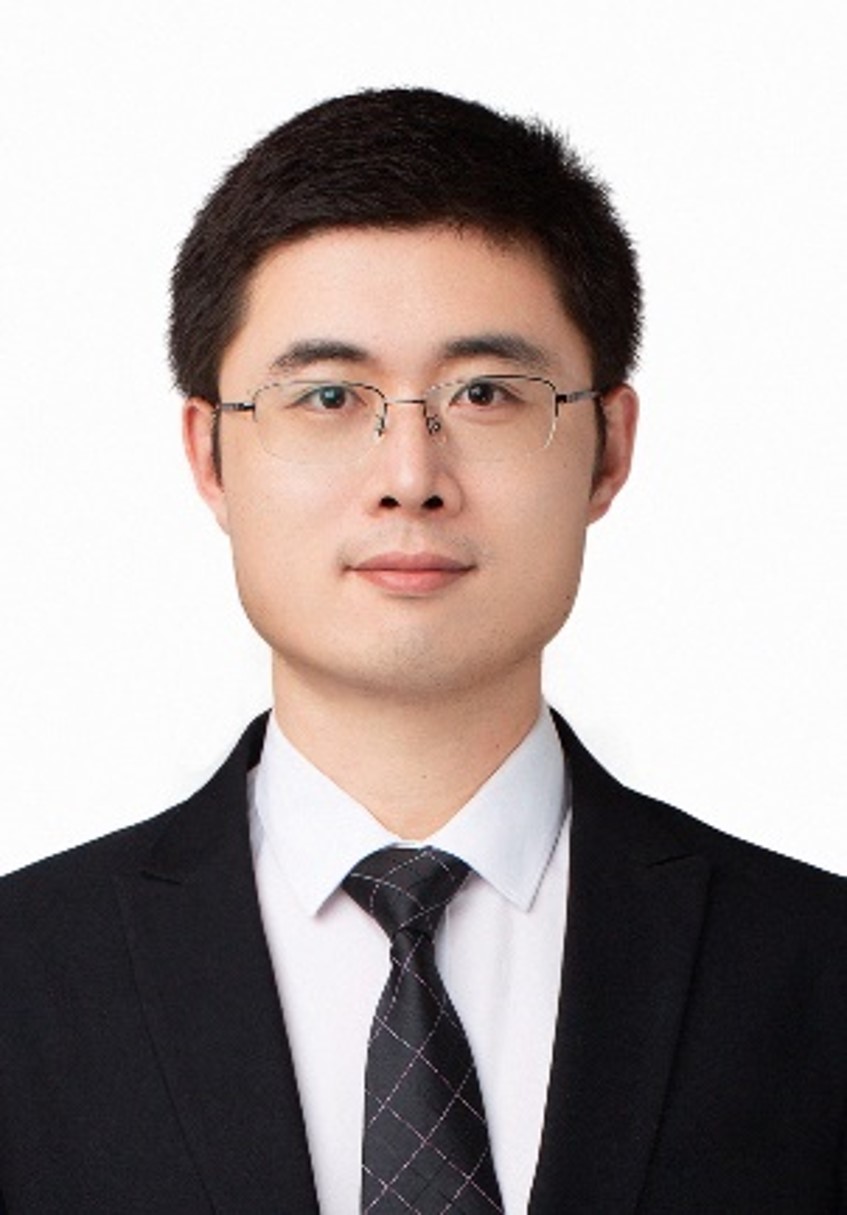}}] 
{Jiankun Wang} (Senior Member, IEEE) received the B.E. degree in automation from Shandong University, Jinan, China, in 2015, and the Ph.D. degree from the Department of Electronic Engineering, The Chinese University of Hong Kong, Hong Kong, in 2019.

During his Ph.D. degree, he spent six months with Stanford University, Stanford, CA, USA, as a Visiting Student Scholar, supervised by Prof. Oussama Khatib. He is currently an Assistant Professor with the Department of Electronic and Electrical Engineering, Southern University of Science and Technology, Shenzhen, China. His current research interests include motion planning and control, human–robot interaction, and machine learning in robotics.
\end{IEEEbiography}

\end{document}